\documentclass[final,5p,times,twocolumn]{elsarticle}

\date{}
\makeatletter
\def\ps@pprintTitle{%
  \let\@oddhead\@empty
  \let\@evenhead\@empty
  \def\@oddfoot{\hfill}%
  \let\@evenfoot\@oddfoot}
\makeatother

\usepackage{amssymb}
\usepackage{amsmath}

\usepackage[figuresright]{rotating}

\usepackage[bookmarks=false]{hyperref}
    \hypersetup{colorlinks,
      linkcolor=blue,
      citecolor=blue,
      urlcolor=blue}

\journal{}

\begin{document}

\begin{frontmatter}



\title{Confidence Adjusted Surprise Measure for Active Resourceful Trials (CA-SMART): A Data-driven Active Learning Framework for Accelerating Material Discovery under Resource Constraints}

\author[label1]{Ahmed Shoyeb Raihan}
\author[label1]{Zhichao Liu}
\author[label2]{Tanveer Hossain Bhuiyan}
\author[label1]{Imtiaz Ahmed}

\affiliation[label1]{organization={Department of Industrial and Management Systems Engineering, West Virginia University},city={Morgantown},postcode={26506},state={West Virginia},country={USA}}

\affiliation[label2]{organization={Department of Mechanical Engineering, The University of Texas at San Antonio}, city={San Antonio},postcode={78249}, state={Texas},country={USA}}

\begin{abstract}
Accelerating the discovery and manufacturing of advanced materials with specific properties is a critical yet formidable challenge due to the vast search space, the high costs of experiments, and the time-intensive nature of material characterization. In recent years, active learning, where a surrogate machine learning (ML) model mimics the scientific discovery process of a human scientist, has emerged as a promising approach to address these challenges by guiding experimentation toward high-value outcomes with a limited budget. Among the diverse active learning philosophies, the concept of surprise—capturing the divergence between expected and observed outcomes—has demonstrated significant potential to drive experimental trials and refine predictive models. Scientific discovery often stems from surprise thereby making it a natural driver to guide the search process. Despite its promise, prior studies leveraging surprise metrics such as Shannon and Bayesian surprise lack mechanisms to account for prior confidence, leading to excessive exploration of uncertain regions that may not yield useful information. To address this, we propose the Confidence-Adjusted Surprise Measure for Active Resourceful Trials (CA-SMART), a novel Bayesian active learning framework tailored for optimizing data-driven experimentation. On a high level, CA-SMART incorporates Confidence-Adjusted Surprise (CAS) to dynamically balance exploration and exploitation by amplifying surprises in regions where the model is more certain while discounting them in highly uncertain areas. This approach aligns with the intuition that prior confidence—whether from a scientist or an ML model—should influence how unexpected outcomes guide future decisions. We evaluated CA-SMART on two benchmark functions (Six-Hump Camelback and Griewank) and in predicting the fatigue strength of steel. The results demonstrate superior accuracy and efficiency compared to traditional surprise metrics, standard Bayesian Optimization (BO) acquisition functions and conventional ML methods. By prioritizing high-impact experimental data points, CA-SMART optimizes decision-making, minimizes resource consumption, and accelerates the discovery of novel materials. The proposed framework establishes a robust foundation for resource-efficient exploration in data-scarce industrial applications, contributing to the broader vision of data-driven decision-making in the era of Industry 5.0.
\end{abstract}




\end{frontmatter}



\section{Introduction}
\label{main}

The rapid discovery of new materials is essential for driving technological advancements, fostering sustainable development, and tackling increasingly complex industrial challenges \cite{stier2024materials}. The demand for novel materials spans diverse fields—from renewable energy and electronics \cite{choubisa2023interpretable} to medicine \cite{pyzer2018bayesian} and aerospace \cite{solomou2018multi}—where advancements hinge on material innovation to improve performance, efficiency, and environmental sustainability \cite{juan2021accelerating}. For instance, breakthroughs in energy storage, such as high-capacity batteries, require materials that combine both high energy density and long cycle life. Similarly, advancements in semiconductors and catalysts depend on the identification of materials with precisely tailored properties that conventional materials cannot provide \cite{cai2020machine}.

Traditional methods for discovering and optimizing materials are often labor-intensive and depend on empirical, trial-and-error approaches \cite{choudhary2022recent, stergiou2023enhancing}. These processes involve synthesizing, testing, and iterating over many candidate materials in laboratory settings, which is both time-consuming and costly \cite{lu2017data}. To overcome these constraints, there has been a significant shift towards leveraging computational tools—especially machine learning (ML) and artificial intelligence (AI)—to accelerate the materials discovery process \cite{fang2022machine, pyzer2022accelerating}. These tools enable researchers to build predictive models based on existing data, offering a faster and more cost-effective means to identify and optimize materials with desired properties \cite{cai2020machine}. For instance, in materials informatics, ML models trained on available datasets can predict the properties of untested materials, effectively narrowing the pool of candidates prior to experimental validation. High-throughput methods have further revolutionized materials science by integrating computational power and automation.  Initiatives like the Materials Genome Initiative \cite{liu2020machine} utilize high-performance computing to simulate thousands or even millions of material candidates, enabling the prediction of structural, electronic, and thermal properties at unprecedented scales \cite{ren2018accelerated}. 

However, the complexity of materials science presents substantial challenges for these approaches. Accurate property prediction requires navigating intricate relationships between a material’s structure, composition, and processing conditions \cite{cole2020design, akinpelu2024discovery}. These factors contribute to a high-dimensional design space, where even minor changes in composition or processing parameters can lead to significant variations in material properties \cite{arroyave2022perspective, bessa2017framework}. Exploring such vast design spaces using traditional experimental methods remains infeasible due to the resource-intensive nature of synthesis and testing \cite{clayson2020high}. Thus, minimizing the number of required trials is critical to optimizing materials efficiently. While ML and high-throughput methods provide significant advantages, they are often constrained in data-scarce environments. High-quality, domain-specific datasets are difficult to obtain, as generating experimental data is both expensive and time-intensive \cite{pilania2021machine, nandy2022audacity}. In such scenarios, traditional ML models may fail to deliver accurate predictions due to their reliance on large, representative datasets \cite{ekins2019exploiting, xu2023small, krishnamurthy2018machine}. This limitation underscores the need for approaches that actively utilize data to iteratively reduce the experimental workload while maintaining high prediction accuracy.

\begin{figure*}[!htb]
    \centering
    \includegraphics[width=\linewidth]{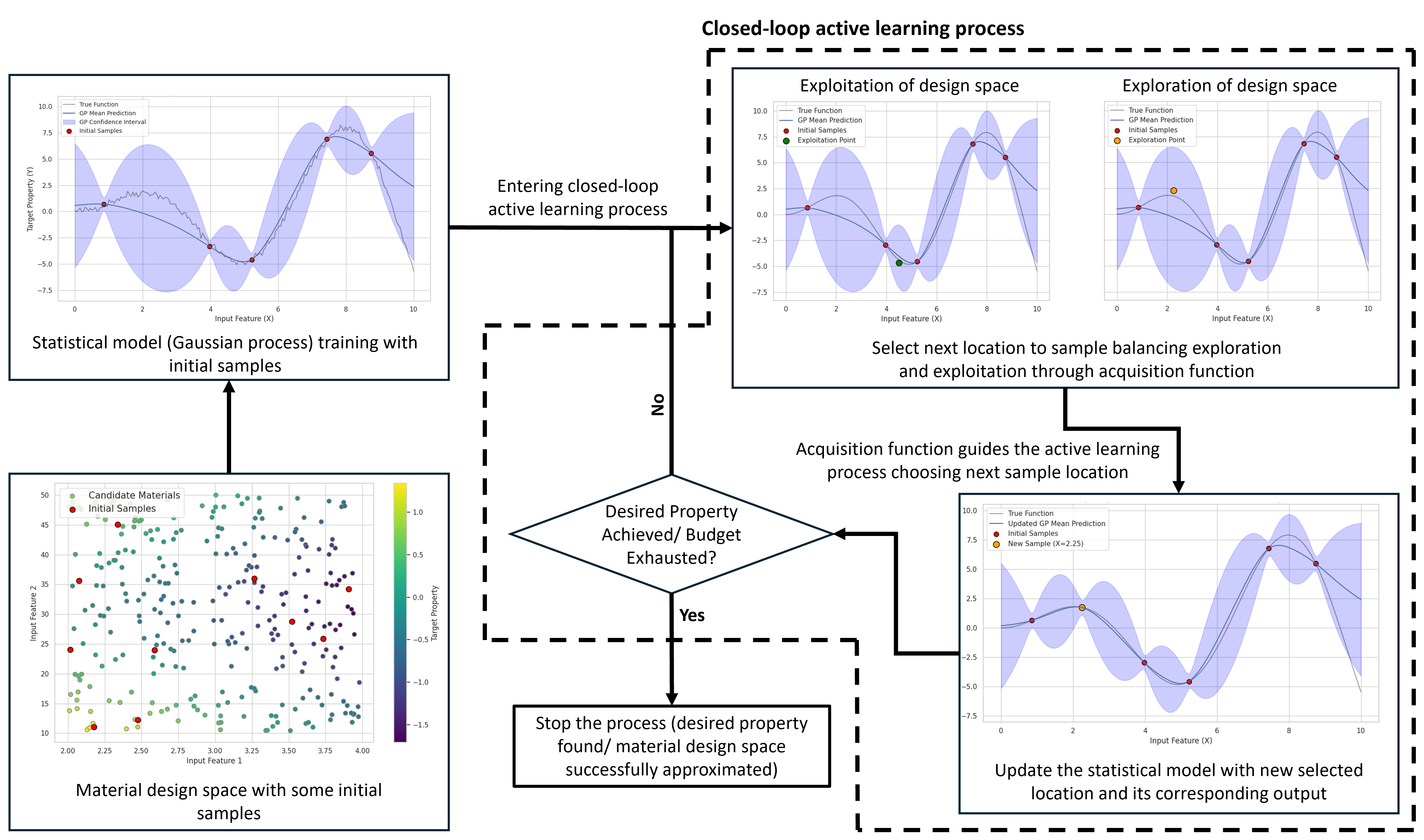}
    \caption{Active learning process in material discovery}
    \label{fig:active_learning_process}
\end{figure*}

Active learning strategies are particularly well-suited for overcoming data limitations in materials science, where generating experimental data is both costly and time-intensive \cite{kusne2020fly}. By iteratively selecting the most informative data points, active learning minimizes the experimental workload while maintaining high prediction accuracy, making it a powerful tool for accelerating material discovery. Figure~\ref{fig:active_learning_process} illustrates the active learning process, which begins with an initial dataset serving as the foundation for training a predictive model. This model generates predictions along with uncertainty estimates, guiding the selection of new sample points through an acquisition function. The acquisition function balances exploitation (targeting regions where the model predicts high performance) and exploration (probing areas of high uncertainty). These selected points are then added to the dataset, and the model is iteratively retrained, progressively improving its predictive accuracy and reducing uncertainty. Bayesian Optimization (BO) is a leading application of the active learning principles described above, designed for optimizing black-box functions that are expensive to evaluate, such as material property prediction \cite{lookman2019active}. BO extends active learning by explicitly formalizing the exploration-exploitation trade-off using a surrogate model, often a Gaussian Process (GP), to approximate the underlying response surface of the black-box function. This surrogate model estimates both the expected value and the uncertainty of predictions, making it well-suited for problems where the true function is unknown and computationally or experimentally expensive to query. BO has been successfully applied across various domains, including materials design \cite{zhang2020bayesian}, additive manufacturing \cite{deneault2021toward}, drug discovery \cite{pyzer2018bayesian}, chemistry \cite{wang2022bayesian}, robotics \cite{burger2020mobile}, and neuroscience \cite{lancaster2018bayesian}.

In BO, the acquisition function identifies the most informative data points by balancing exploration and exploitation. This structured approach maximizes information gain from each experiment, enabling rapid convergence toward optimal material candidates with minimal experimental trials \cite{zuo2021accelerating}. However, traditional acquisition functions in BO, such as Expected Improvement (EI) \cite{jones1998efficient} and Upper Confidence Bound (UCB) \cite{snoek2012practical}, tend to lean too heavily toward either exploration or exploitation \cite{raihan2024accelerating}. For instance, EI is often more exploitative, focusing primarily on promising areas and potentially overlooking novel regions that may yield higher overall performance. Conversely, UCB favors exploration by selecting points with high uncertainty, which can sometimes lead to unnecessary trials in regions with low likelihoods of improvement. Balancing exploration and exploitation thus remains a fundamental challenge in active learning and BO frameworks. This is critical for material discovery and similar fields, where the aim is to achieve optimal predictions with minimal experimental trials. 

Therefore, instead of relying on a predefined static balance between these two strategies, the acquisition process should be adaptive, driven by changes in belief brought by the new data point. Consequently, recent research has explored the concept of `surprise' as an alternative, drawing inspiration from neuroscience and cognitive psychology \cite{modirshanechi2022taxonomy}. This notion, formalized through different metrics in the literature, quantifies how much an observation challenges current beliefs \cite{liakoni2022brain, faraji2018balancing}. Surprise-based active learning processes address the drawbacks of classical BO by introducing a mechanism to guide sampling toward regions that yield the highest information gain \cite{avila2014behavior}. Unlike traditional acquisition functions that rely solely on statistical properties such as mean or variance, surprise-based methods quantify the divergence between the model’s prior expectations and observed outcomes \cite{jin2022autonomous, ahmed2024toward}. This enables the framework to adaptively shift focus between exploitation of well-understood regions and exploration of uncharted areas, dynamically balancing the trade-off without relying on predefined hyperparameters. Furthermore, the integration of surprise measures ensures that each experiment maximizes its contribution to refining the surrogate model, enhancing data efficiency, and reducing the number of iterations needed to achieve convergence \cite{raihan2024augmented}. 

However, existing surprise measures, such as Shannon and Bayesian surprise, are not perfect and inadequately address the nuances of uncertainty in the BO process \cite{ahmed2024toward}. In this work, we propose a Confidence-Adjusted Surprise Measure for Active Resourceful Trials (CA-SMART), a novel Bayesian active learning framework. At its core, CA-SMART employs the Confidence-Adjusted Surprise (CAS) measure, which uniquely integrates the magnitude of surprise with the model's confidence in a given region. While Shannon surprise quantifies the unexpectedness of an observation based on its likelihood, it often overemphasizes rare events without considering their relevance or credibility \cite{shannon1948mathematical, barto2013novelty}. Similarly, Bayesian surprise captures the divergence between prior and posterior beliefs, emphasizing changes in understanding but neglecting the reliability of the observations themselves \cite{schmidhuber2010formal, mousavi2022spatiotemporal}. CAS addresses these limitations by incorporating model confidence into the evaluation of surprise. This integration allows CAS to amplify the impact of surprising outcomes in regions where the model demonstrates high certainty while discounting surprises in highly uncertain areas, where observations are less credible. By doing so, CAS prevents the framework from over-exploring uncertain regions that are unlikely to yield actionable insights, striking a dynamic balance between exploration and exploitation. The inclusion of confidence is pivotal in guiding CA-SMART’s adaptive search process. It ensures that model updates are driven by observations that are both unexpected and reliable, effectively refining predictions through high-impact experiments \cite{faraji2018balancing}.  We have validated CA-SMART on two synthetic benchmark functions (Six-Hump Camelback and Griewank) demonstrating its ability to approximate these complex synthetic functions with fewer iterations compared to traditional acquisition functions and existing surprise-based approaches in the BO literature. Following these validations, we have also conducted a case study focused on predicting the fatigue strength of steel, using data sourced from the Japan National Institute of Materials (NIMS) fatigue strength dataset, which includes comprehensive information on composition, heat treatment, and inclusion parameters \cite{agrawal2016fatigue}. CA-SMART has exhibited superior accuracy and data efficiency in predicting fatigue strength of steel, underscoring its potential for resource-constrained, data-scarce domains such as material discovery.

The rest of this paper is organized as follows. Section 2 provides a comprehensive review of related work, covering traditional, machine learning, and active learning-based approaches for material discovery highlighting the need for more adaptive approaches. Section 3 describes the theoretical foundations of BO, provides background on existing surprise measures, and introduces the proposed CA-SMART framework, detailing its methodology, including the CAS measure and its implementation within the BO framework. In Section 4, we present the experimental results, beginning with evaluations on synthetic benchmark functions, followed by a case study on predicting the fatigue strength of steel. Comparative analyses with other surprise-based methods and traditional BO acquisition functions are provided, using RMSE and CRPS as performance metrics. Section 5 concludes the paper by summarizing key findings, highlighting the advantages of CA-SMART, and outlining potential directions for future research.

\section{Related Works}

\subsection{Traditional Approaches to Material Discovery}

The traditional process of discovering new materials has largely been a slow and labor-intensive endeavor, often spanning several years from initial research to practical application \cite{papadimitriou2024ai, fang2022machine}. This prolonged timeline is due to the reliance on repetitive experimental and theoretical characterization studies, which require a combination of chemical intuition and serendipity \cite{liu2017materials}. Historically, two conventional methods—experimental measurement and computational simulation—have dominated the field of materials science \cite{schleder2019dft}. Experimental measurements, encompassing microstructure and property analysis, synthetic experiments, and property measurements, have been the cornerstone of materials research due to their intuitive and direct nature \cite{liu2017materials}. However, these approaches are inherently time-consuming, resource-intensive, and require specialized equipment, controlled environments, and expert researchers \cite{brunton2019methods}. Furthermore, as the complexity of material systems increases, experimental techniques alone struggle to adequately capture the intrinsic relationships between material characteristics and their properties \cite{fang2022machine}.

To complement experimental methods, computational simulations have become an indispensable tool in materials research \cite{louie2021discovering}. Techniques such as Density Functional Theory (DFT)-based electronic structure calculations \cite{sholl2022density}, molecular dynamics \cite{mahmood2022machine}, Monte Carlo simulations \cite{singh2023basics}, phase-field modeling \cite{zhao2023understanding}, and finite element analysis \cite{peirlinck2024automated} have significantly enhanced the ability to model and predict material behaviors. These methods provide a controlled and cost-effective alternative to physical experimentation, enabling researchers to virtually conduct experiments and explore the impact of various parameters on material properties. For instance, modern computational tools have the potential to reduce the material development timeline from decades to as little as 18 months \cite{hautier2012computer}. However, despite their promise, these methods face significant limitations. They often demand high-performance computing resources and rely heavily on the microstructure and specific characteristics of the materials being studied \cite{liu2017materials}. Additionally, the lack of reusability of computational results across diverse material systems hampers broader applicability and scalability. Addressing these challenges—particularly the increasing complexity of material systems and the demand for accelerated discovery—calls for the adoption of ML-driven approaches that leverage data reusability, resource efficiency, and scalability, setting the stage for transformative advancements in materials discovery.

\subsection{Machine Learning Approaches to Material Discovery}

In recent years, the field of materials science has experienced a paradigm shift with the integration of ML methods \cite{huang2022machine, vivanco2022machine}. ML enables researchers to uncover patterns and relationships in high-dimensional datasets without relying solely on explicit physical models \cite{murphy2012machine, fang2022machine}. This capability has positioned ML as a core technology in the emerging field of materials informatics. A variety of ML algorithms have been applied in materials science, each tailored to specific tasks such as property prediction \cite{chibani2020machine}, material classification \cite{penumuru2020identification}, defect identification \cite{fu2022machine}, process optimization \cite{liu2015predictive}, and discovery of novel materials \cite{raihan2024accelerating}. Among these techniques, k-Nearest Neighbors (KNN) serves as an intuitive, proximity-based classification and regression approach. KNN has been successfully employed in predicting phases in high-entropy alloys (HEAs), demonstrating robust performance in phase prediction tasks due to its simplicity and adaptability in data-rich environments \cite{ghouchan2022graph}. However, KNN often struggles with scalability and performance in large, noisy datasets. To overcome these challenges, Support Vector Machines (SVMs) and Support Vector Regression (SVR) employ hyperplanes to separate data points in high-dimensional space, making them more suitable for complex, nonlinear relationships. For instance, SVR has been effectively applied to predict the indirect tensile strength (ITS) of foamed bitumen-stabilized base course materials, outperforming simpler models due to its kernel-based approach, which enables high accuracy even with limited data \cite{nazemi2016support}. While SVMs/SVRs deliver strong performance, their computational demands can become significant for large datasets. Artificial Neural Networks (ANNs) further extend the capabilities of ML by mimicking the structure and function of biological neural networks, allowing them to model highly nonlinear relationships. ANNs have been widely applied in materials science, particularly in predicting the mechanical and tribological properties of composite materials \cite{paturi2022role}. However, ANNs require substantial data and computational resources for effective training and are prone to overfitting without proper regularization.

Random Forests (RF) offer an ensemble-based solution to mitigate overfitting and improve robustness. By averaging predictions from multiple decision trees trained on bootstrapped datasets, RF models achieve higher accuracy and reliability compared to SVMs or standalone decision trees. For example, RF has been successfully applied in material classification tasks, such as identifying and discriminating grades of iron ore based on their chemical compositions \cite{sheng2015classification}. Gradient Boosting Machines (GBMs) take ensemble learning a step further by sequentially training decision trees to minimize errors iteratively. GBMs excel in capturing complex, non-linear interactions between variables, as demonstrated in predicting the compressive strength of high-performance concrete (HPC). Their ability to balance bias and variance makes them particularly suitable for regression and classification tasks in materials science \cite{kaloop2020compressive}.

As the complexity of material data increases, Deep Neural Networks (DNNs) have emerged as powerful tools for representation learning. Their hierarchical feature extraction capabilities enable them to model intricate patterns in high-dimensional datasets effectively. For instance, DNNs have been applied to predict elastic properties of materials using three-dimensional electronic charge density data, achieving superior accuracy and adaptability to diverse material systems \cite{zhao2020predicting}. Despite their strengths, DNNs require large datasets and significant computational resources, which can limit their applicability in data-scarce scenarios. Finally, Gaussian Process Regression (GPR) stands out for its probabilistic framework, providing both predictions and uncertainty quantifications. GPR has been shown to outperform other ML techniques, including SVR, RF, and DNNs, in predicting the remaining fatigue life of metallic materials under two-step loading \cite{gao2022gaussian}. Its ability to handle small datasets and quantify uncertainty makes it an invaluable tool for materials discovery in noisy or data-constrained environments \cite{farid2022data}. However, when trained on static data without sequential updates, GPR may face similar limitations as other regression techniques, such as reduced adaptability to evolving systems.

\subsection{Active Learning Approaches to Material Discovery}

While ML approaches have revolutionized material discovery by utilizing data-driven techniques to predict properties, classify materials, and optimize processes, their success often hinges on the availability of large, high-quality datasets. However, acquiring such data in material science can be prohibitively expensive and time-consuming due to the intricate experimental and computational procedures involved. Active learning addresses this challenge by strategically focusing experimental efforts on the most informative data points, thereby reducing the number of trials required to achieve accurate predictions \cite{raihan2024accelerating}.  Active learning techniques can be broadly classified into two prominent categories based on their underlying frameworks: Reinforcement Learning (RL)-based approaches and Bayesian Optimization (BO)-based approaches.

\subsubsection{Reinforcement Learning (RL)-based Approaches}

Reinforcement Learning (RL), a branch of sequential ML, excels at addressing sequential decision-making problems under uncertainty, making it highly suitable for applications in materials science. RL agents iteratively interact with their environment, learning optimal strategies by maximizing cumulative rewards, which is particularly beneficial for solving complex, high-dimensional tasks such as material and discovery \cite{gow2022review}. For instance, RL has been used to optimize multi-step chemical synthesis processes in self-driving labs, as demonstrated by the AlphaFlow system, which autonomously identified and optimized synthetic routes for core-shell semiconductor nanoparticles, navigating a 40-dimensional parameter space \cite{volk2023alphaflow}. Similarly, RL has proven effective in predicting optimal chemical vapor deposition (CVD) schedules for synthesizing semiconducting MoS$_2$, leveraging simulated data to identify time-dependent reaction conditions that enhance crystallinity and phase purity while minimizing resource use \cite{rajak2021autonomous}. Building on these advances, the ReLMM framework utilizes a multi-agent RL approach to optimize feature selection, identifying minimal yet informative feature sets for semiconducting materials with superior accuracy \cite{sharma2024relmm}. Similarly, RL is applied to materials microstructure optimization and found to surpass traditional combinatorial approaches \cite{vasudevan2022discovering}.

Despite the promise, its broader adoption in materials discovery remains at an early stage. The limitations stem from several challenges inherent to RL. First, RL methods often require extensive interaction with the environment, which can be computationally prohibitive or experimentally unfeasible. Second, RL agents typically struggle with noisy, sparse, or incomplete data, which are common in real-world materials datasets. Lastly, the high-dimensional and heterogeneous nature of material design spaces makes training RL models difficult without extensive tuning or domain-specific adjustments. This underlines the need for alternative methods that emphasize efficient exploration and optimization in high-dimensional, resource-intensive settings

\subsubsection{Bayesian Optimization (BO)-based Approaches}

BO has emerged as a powerful sequential ML framework in materials discovery, excelling in optimizing expensive black-box functions through sequential data acquisition and surrogate modeling \cite{wang2022bayesian}. Its iterative approach is particularly suited for the high-dimensional and resource-intensive challenges of materials science. BO has demonstrated success in various material science applications \cite{deshwal2021bayesian, yamashita2018crystal, chitturi2024targeted}. Recent advancements, such as mixed-variable BO using latent-variable GPs (LVGP), enable simultaneous optimization of qualitative and quantitative variables, driving breakthroughs in solar cell absorption and hybrid perovskite design \cite{zhang2020bayesian}.

Traditionally, GP has been the primary surrogate model for BO, valued for its ability to capture uncertainty and provide robust predictions. \cite{herbol2018efficient, diwale2022bayesian}. Beyond regular GPs, alternative surrogate models have expanded BO's applicability. Anisotropic GPs have shown superior robustness compared to isotropic versions, while random forest (RF)-based BO models offer computational efficiency, requiring less hyperparameter tuning and being less sensitive to data distribution \cite{liang2021benchmarking}. Adaptive models, such as Bayesian multivariate adaptive regression splines and Bayesian additive regression trees, have further enhanced BO's ability to handle high-dimensional, non-smooth objective functions \cite{lei2021bayesian}. The PAL 2.0 framework integrates physics-based priors with advanced machine learning models like XGBoost and neural networks to improve BO's efficiency in materials discovery, including perovskites and organic thermoelectric semiconductors \cite{priyadarshini2024pal}. 

In BO, the success of the framework hinges on balancing exploration and exploitation, particularly in resource-constrained and high-dimensional scenarios common in materials discovery. Traditional acquisition functions, as discussed in Section 1, such as Expected Improvement (EI) \cite{jones1998efficient} and Upper Confidence Bound (UCB) \cite{snoek2012practical}, often struggle to maintain this balance, leading to inefficiencies in optimization processes where experiments are costly and data is scarce \cite{raihan2024accelerating}. While recent advancements, such as surprise-based approaches inspired by Shannon and Bayesian surprise metrics \cite{modirshanechi2022taxonomy, ahmed2024toward}, offer promising adaptive strategies, challenges remain in effectively quantifying uncertainty and dynamically managing the trade-off. Addressing these limitations is crucial for enhancing data efficiency and accuracy in identifying optimal material candidates, emphasizing the need for innovative acquisition mechanisms tailored to materials discovery \cite{pyzer2018bayesian, burger2020mobile}.

\section{Research Methodology}

In this section, we present the theoretical and methodological foundation underlying our active learning framework, CA-SMART (Confidence-Adjusted Surprise Measure for Active Resourceful Trials). We begin with an overview of BO, followed by an introduction to the concept of surprise in active learning, covering traditional metrics—Shannon surprise and Bayesian surprise—and their roles in balancing exploration and exploitation. Building on these, we introduce Confidence-Adjusted Surprise which we propose to use as an acquisition function. Finally, we describe how CA-SMART utilizes CAS to improve material property prediction with minimal experimental trials.

\subsection{Bayesian Optimization}

BO is a sequential, model-based approach widely used for optimizing expensive black-box functions. It is particularly useful in high-dimensional search spaces where direct evaluations of the objective function are costly, as in materials science, drug discovery, and robotics \cite{frazier2016bayesian}. BO combines a probabilistic surrogate model, typically a GP, with an acquisition function to guide the selection of new data points. This surrogate model provides an estimate of the objective function and a measure of uncertainty, which is essential for balancing exploration (evaluating uncertain regions) and exploitation (focusing on promising regions) in the search space. In BO, the GP model assumes that any finite number of function observations have a joint Gaussian distribution, with a mean function and a covariance function defining the relationship between points \cite{brochu2010tutorial}. As new observations are added, the GP is updated, refining its predictions and uncertainties across the domain. The acquisition function, such as EI or UCB, then identifies the next point to evaluate by maximizing a criterion that balances exploration and exploitation, thereby efficiently navigating the search space with minimal evaluations.

\subsubsection{Gaussian Process as a Surrogate Model}

Gaussian Processes (GPs) provide a powerful, non-parametric approach for modeling complex, high-dimensional functions. In BO, GPs are widely used as surrogate models to approximate the underlying objective function \(f(\mathbf{x})\), enabling efficient prediction and uncertainty quantification in unexplored regions of the search space \cite{rasmussen2003gaussian}.

A GP is defined as a collection of random variables, any finite subset of which has a joint Gaussian distribution. For an input \( \mathbf{x} \in \mathbb{R}^d \), a GP defines a distribution over possible functions \( f(\mathbf{x}) \) that are consistent with prior observations. The GP is characterized by a mean function \( m(\mathbf{x}) \) and a covariance function \( k(\mathbf{x}, \mathbf{x}') \), also known as the kernel:

\begin{equation}
f(\mathbf{x}) \sim \mathcal{GP}(m(\mathbf{x}), k(\mathbf{x}, \mathbf{x}'))
\end{equation}

where \( m(\mathbf{x}) = \mathbb{E}[f(\mathbf{x})] \) is the mean function, often assumed to be zero for simplicity, and \( k(\mathbf{x}, \mathbf{x}') = \mathbb{E}[(f(\mathbf{x}) - m(\mathbf{x}))(f(\mathbf{x}') - m(\mathbf{x}'))] \) is the covariance function, which defines the smoothness and generalization properties of the modeled function.

Given a set of observed data points \( \mathcal{D} = \{\mathbf{X}, \mathbf{y}\} \), where \( \mathbf{X} = \{\mathbf{x}_1, \mathbf{x}_2, \ldots, \mathbf{x}_n\} \) and \( \mathbf{y} = \{y_1, y_2, \ldots, y_n\} \), we model the joint distribution of the observed outputs \( \mathbf{y} \) and the function value at a new test point \( \mathbf{x}_* \) as:

\begin{equation}
\begin{bmatrix}
\mathbf{y} \\
f(\mathbf{x}_*)
\end{bmatrix} \sim \mathcal{N} \left(
\begin{bmatrix}
\mathbf{0} \\
0
\end{bmatrix},
\begin{bmatrix}
\mathbf{K}(\mathbf{X}, \mathbf{X}) + \sigma_n^2 \mathbf{I} & \mathbf{k}(\mathbf{X}, \mathbf{x}_*) \\
\mathbf{k}(\mathbf{x}_*, \mathbf{X}) & k(\mathbf{x}_*, \mathbf{x}_*)
\end{bmatrix}
\right)
\end{equation}

where \( \mathbf{K}(\mathbf{X}, \mathbf{X}) \) is the covariance matrix of observed points, \( \mathbf{k}(\mathbf{X}, \mathbf{x}_*) \) represents the covariance vector between the test point \( \mathbf{x}_* \) and training points in \( \mathbf{X} \), and \( \sigma_n^2 \) is the observation noise variance. The predictive posterior mean and variance at \( \mathbf{x}_* \) are then given by:

\begin{equation}
\begin{array}{l}
\mathbb{E}[f(\mathbf{x}_*) | \mathcal{D}] = \mathbf{k}(\mathbf{X}, \mathbf{x}_*)^\top \left(\mathbf{K}(\mathbf{X}, \mathbf{X}) + \sigma_n^2 \mathbf{I} \right)^{-1} \mathbf{y}, \\[10pt]
\text{Var}[f(\mathbf{x}_*) | \mathcal{D}] = k(\mathbf{x}_*, \mathbf{x}_*) \\
\qquad - \mathbf{k}(\mathbf{X}, \mathbf{x}_*)^\top \left(\mathbf{K}(\mathbf{X}, \mathbf{X}) + \sigma_n^2 \mathbf{I} \right)^{-1} \mathbf{k}(\mathbf{X}, \mathbf{x}_*)
\end{array}
\end{equation}

The choice of the covariance function \( k(\mathbf{x}, \mathbf{x}') \), or kernel, plays a crucial role in determining the performance of the GP model. Three widely used kernels are discussed in the following. The Radial Basis Function (RBF) kernel, also known as the squared exponential kernel, assumes smooth and infinitely differentiable functions. It is defined as:

\begin{equation}
k_{\text{RBF}}(\mathbf{x}, \mathbf{x}') = \theta^2 \exp \left( -\frac{\|\mathbf{x} - \mathbf{x}'\|^2}{2\ell^2} \right)
\end{equation}

where \( \theta \) is the signal variance and \( \ell \) is the length scale, controlling the smoothness of the function. The Matern kernel is more flexible than the RBF kernel and can model functions that are less smooth. It is defined as:

\begin{equation}
k_{\text{Matern}}(\mathbf{x}, \mathbf{x}') = \theta^2 \frac{2^{1-\nu}}{\Gamma(\nu)} \left( \frac{\sqrt{2\nu} \|\mathbf{x} - \mathbf{x}'\|}{\ell} \right)^\nu K_\nu \left( \frac{\sqrt{2\nu} \|\mathbf{x} - \mathbf{x}'\|}{\ell} \right)
\end{equation}

where \( \nu \) is a smoothness parameter, \( K_\nu \) is the modified Bessel function, and \( \ell \) is the length scale. The commonly used values of \( \nu \) include \( \nu = 1.5 \) (once differentiable) and \( \nu = 2.5 \) (twice differentiable). The Rational Quadratic kernel is a scale mixture of RBF kernels with different length scales, making it suitable for capturing both large and small variations in the data. It is given by:

\begin{equation}
k_{\text{RQ}}(\mathbf{x}, \mathbf{x}') = \theta^2 \left( 1 + \frac{\|\mathbf{x} - \mathbf{x}'\|^2}{2\alpha \ell^2} \right)^{-\alpha}
\end{equation}

where \( \alpha \) controls the relative weighting of length scales, \( \ell \) is the length scale, and \( \theta \) is the signal variance. The selection of an appropriate kernel depends on the characteristics of the objective function. 

\subsubsection{Acquisition Functions}
Acquisition functions play a critical role in BO by guiding the selection of new sampling points based on the surrogate model's predictions \cite{brochu2010tutorial}. By balancing exploration and exploitation, acquisition functions help identify optimal points efficiently while minimizing the number of costly evaluations. Several acquisition functions are commonly used in BO, including Expected Improvement (EI) \cite{jones1998efficient}, Probability of Improvement (PI) \cite{brochu2010tutorial}, Upper Confidence Bound (UCB) \cite{snoek2012practical}, Knowledge Gradient (KG) \cite{frazier2018bayesian}, and Entropy Search (ES) \cite{frazier2016bayesian}. Each of these functions offers a unique approach to balancing exploration and exploitation, tailored to specific optimization goals. Among them, the EI acquisition function is one of the most widely used, focusing on maximizing the expected improvement over the best observed value. Given the best observed value \( f(\mathbf{x}_{\text{best}}) \) and the Gaussian Process model predictions, the EI function at a candidate point \( \mathbf{x}_* \) is defined as:

\begin{equation}
    \text{EI}(\mathbf{x}_*) = \mathbb{E}\left[\max(0, f(\mathbf{x}_*) - f(\mathbf{x}_{\text{best}}))\right]
\end{equation}

Since \( f(\mathbf{x}_*) \) is normally distributed with mean \( \mu(\mathbf{x}_*) \) and variance \( \sigma^2(\mathbf{x}_*) \), the EI function can be expressed as:

\begin{equation}
    \text{EI}(\mathbf{x}_*) = (\mu(\mathbf{x}_*) - f(\mathbf{x}_{\text{best}})) \Phi(Z) + \sigma(\mathbf{x}_*) \phi(Z)
\end{equation}

where \( Z = \frac{\mu(\mathbf{x}_*) - f(\mathbf{x}_{\text{best}})}{\sigma(\mathbf{x}_*)} \), \( \Phi(Z) \) is the cumulative distribution function of the standard normal distribution, and \( \phi(Z) \) is the corresponding probability density function. EI is effective in balancing exploration and exploitation by assigning high values to points with potential improvements over the current best observation.

\begin{figure*}[!htb]
    \centering
    \includegraphics[width=\linewidth]{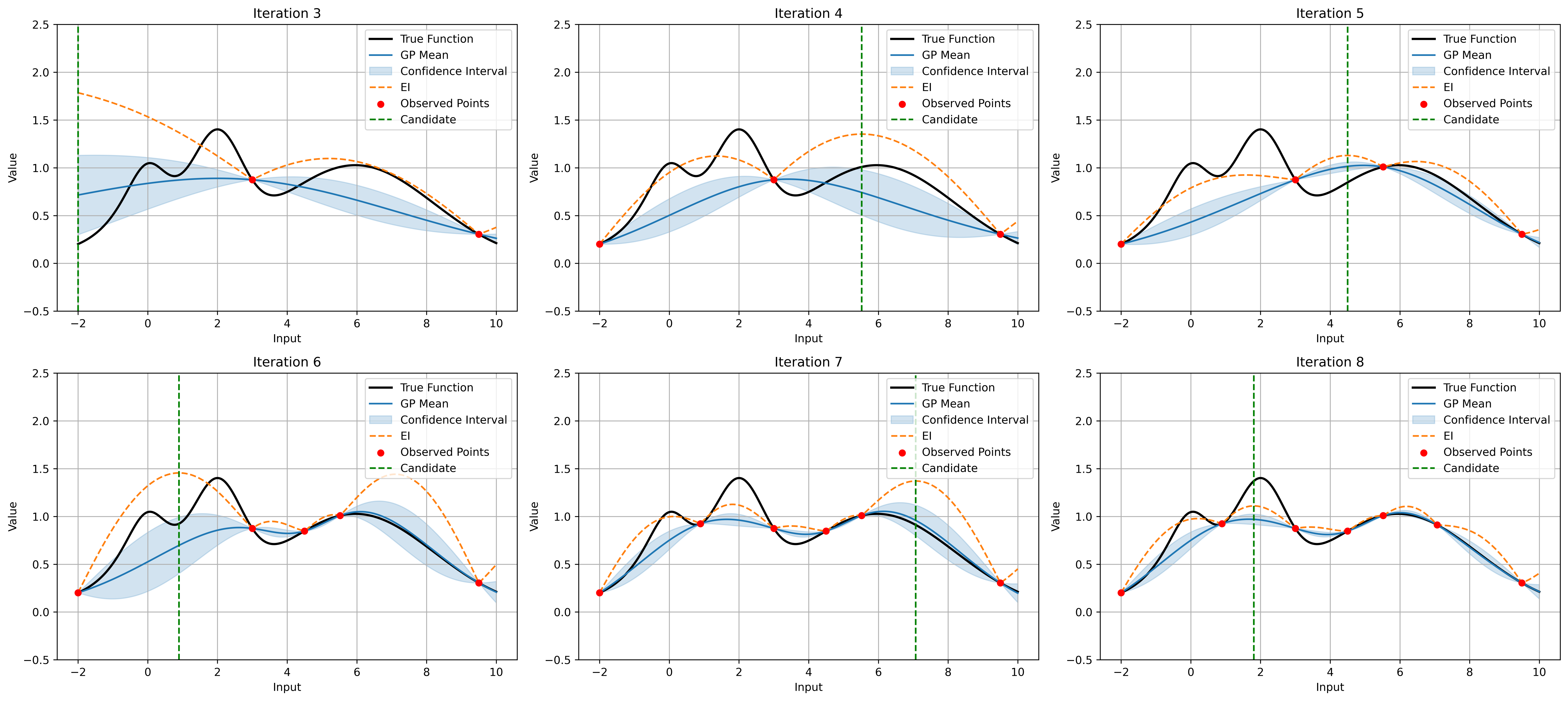}
    \caption{Bayesian Optimization process over multiple iterations}
    \label{fig:bo_iterations}
\end{figure*}

Figure \ref{fig:bo_iterations} demonstrates the BO process over multiple iterations for a one-dimensional function, showcasing the adaptive sampling mechanism. In each iteration, the black solid line represents the true underlying function, while the blue line indicates the GP model’s predicted mean. The shaded blue region corresponds to the GP’s confidence interval, providing an estimate of uncertainty in the model’s predictions. The red dots highlight the observed data points up to that iteration, and the orange dashed line represents the EI acquisition function, which is used to balance exploration and exploitation. At each iteration, the green vertical line marks the candidate point selected by the EI acquisition function, which aims to maximize the potential for informative sampling. As the iterations progress, the GP model increasingly approximates the true function more accurately, reducing uncertainty in critical areas. This iterative approach illustrates how BO efficiently narrows down the search space to regions with high information gain, thus optimizing the function with minimal evaluations.

\subsection{Surprise Measures in Active Learning}

The concept of surprise has recently gained prominence as a dynamic and adaptive strategy in active learning and BO, addressing some of the limitations of traditional acquisition functions \cite{raihan2024augmented, ahmed2024toward}. Borrowed from fields such as neuroscience and cognitive psychology, surprise quantifies the discrepancy between observed outcomes and the expectations formed by a predictive model \cite{ahmed2024toward}. Neuroscience research suggests that surprise acts as a cognitive trigger, encouraging adaptive responses and deeper exploration of new information \cite{modirshanechi2022taxonomy}. In the context of active learning, surprise provides a metric to identify observations that deviate significantly from a model's predictions, thereby guiding experimentation toward regions of the search space that offer high potential for new insights \cite{lancaster2018bayesian}. Surprise can be viewed through three distinct philosophical approaches, each reflecting a different type of discrepancy between predictions and observations: probabilistic mismatch, observation mismatch, and belief mismatch \cite{modirshanechi2022taxonomy}. We will now discuss various surprise metrics available in the literature that align with these approaches.


\subsubsection{Shannon Surprise}

Shannon surprise is rooted in information theory and provides a measure of the unexpectedness of an observation relative to the model's predicted probability distribution. It falls under the category of probabilistic mismatch surprise, focusing on low-probability events that deviate significantly from the norm. By quantifying the information content of an observation, Shannon surprise identifies data points that significantly deviate from the current predictive model, emphasizing regions of the search space where the model's uncertainty is high or its predictions are less confident \cite{baldi2002computational}. Mathematically, Shannon surprise is defined as:

\begin{equation}
S_{\text{Shannon}} = -\log p(y | \mathbf{x}, \mathcal{M})
\end{equation}

where \( p(y | \mathbf{x}, \mathcal{M}) \) represents the likelihood of the observed value \( y \) given the input \( \mathbf{x} \) and the model \( \mathcal{M} \), characterized by its predictive distribution. The negative logarithm ensures that less likely observations correspond to higher values of \( S_{\text{Shannon}} \), effectively quantifying the unexpectedness of an observation. When an observation lies in the tail of the predicted distribution, its likelihood is low, resulting in a high surprise value. This property makes Shannon surprise particularly effective for anomaly detection and identifying unexplored regions of the search space. For instance, in a GP model predicting the electrical conductivity of alloys, the predictive distribution is characterized by a mean \( \mu(\mathbf{x}) \) and variance \( \sigma^2(\mathbf{x}) \). If a new observation lies far from the predicted mean (e.g., more than three standard deviations away), the likelihood \( p(y | \mathbf{x}, \mathcal{M}) \) will be low, producing a high \( S_{\text{Shannon}} \) value. Such observations are flagged as anomalies, prompting further investigation into whether the deviation stems from experimental error, novel material behavior, or unaccounted factors.

\subsubsection{Bayesian Surprise}

Bayesian surprise extends the concept of unexpectedness in an observation by focusing on how a new observation modifies the model's internal belief state. Unlike Shannon surprise, which evaluates the likelihood of an observation under the prior predictive distribution, Bayesian surprise quantifies the degree to which a single observation causes the model's understanding to shift. This shift is mathematically captured using the KL divergence between the prior and posterior distributions over the model’s parameters, making Bayesian surprise a belief mismatch metric \cite{itti2009bayesian, faraji2018balancing}. Mathematically, Bayesian surprise is expressed as:

\begin{align}
S_{\text{Bayesian}} &= \mathrm{KL}\Bigl( p(\mathcal{M}_{\text{posterior}}) \,\Big\|\, p(\mathcal{M}_{\text{prior}}) \Bigr) \\
                   &= \mathrm{KL}\Bigl( p(\mathcal{M} \mid D) \,\Big\|\, p(\mathcal{M}) \Bigr) \nonumber \\
                   &= \int p(\mathcal{M} \mid D) \, \log \frac{p(\mathcal{M} \mid D)}{p(\mathcal{M})} \, d\mathcal{M}\nonumber
\end{align}

where, \( D = \{\mathbf{x}, y\} \) denotes the new data point, with \( \mathbf{x} \) as the input vector and \( y \) as the scalar output, \( p(\mathcal{M}_{\text{prior}}) \) represents the prior distribution over the model parameters before observing \( D \), and \( p(\mathcal{M}_{\text{posterior}}) \) represents the posterior distribution over the model parameters after incorporating the observation \( D \). The KL divergence quantifies the information gain from an observation \( D \) by measuring how much the posterior distribution \( p(\mathcal{M}_{\text{posterior}}) \) diverges from the prior distribution \( p(\mathcal{M}_{\text{prior}}) \). For Gaussian distributions, Bayesian surprise can be explicitly calculated as:

\begin{equation}
S_{\text{Bayesian}} = \log \frac{\sigma_{\text{prior}}}{\sigma_{\text{posterior}}} + 
\frac{\sigma_{\text{posterior}}^2 + (\mu_{\text{posterior}} - \mu_{\text{prior}})^2}{2\sigma_{\text{prior}}^2} - 0.5
\end{equation}

where \( \mu_{\text{prior}} \), \( \sigma_{\text{prior}} \), \( \mu_{\text{posterior}} \), and \( \sigma_{\text{posterior}} \) represent the mean and variance of the prior and posterior predictive distributions, respectively. Bayesian surprise measures the extent to which an observation \( D \) updates the model’s beliefs, highlighting data points that provide significant insights into the system. High surprise values occur when an observation substantially shifts the prior distribution, indicating a meaningful gain in information, whereas low values reflect minimal belief updates. This mechanism is particularly useful for identifying observations that challenge the model's assumptions or reveal novel relationships in the data.

\subsection{Confidence-Adjusted Surprise (CAS)}

The CAS is a robust and adaptive metric that integrates probabilistic measures of surprise, belief updates, and corrections for model confidence to identify observations that are both surprising and informative. CAS incorporates Shannon surprise, which quantifies the unexpectedness of an observation based on the model’s predictive distribution, and Bayesian surprise, which measures the shift in model belief after an observation. Additionally, CAS introduces a confidence correction term to account for predictive uncertainty and an adjustment term to align the framework with a flat prior baseline. These components work together to enable CAS to dynamically refine the model’s understanding of the search space as new observations are made. Given an observation \( \mathbf{D} = \{\mathbf{x}, y\} \), CAS is defined as:

\begin{equation}
S_{\text{Confidence-Adjusted}} = S_{\text{Shannon}} + S_{\text{Bayesian}_\text{flat}} + C - A
\end{equation}

where \( S_{\text{Shannon}} \) quantifies the unexpectedness of the observed response \( y \) based on the GP model's current predictive distribution, \( S_{\text{Bayesian}_\text{flat}} \) measures the shift in model belief after incorporating the observation \( D \), \( C \) adjusts for the model's confidence in its predictions, and \( A \) aligns the CAS with a flat (uninformative) prior to ensure robustness. It is important to note that, here, we use \( S_{\text{Bayesian}_\text{flat}} \), which is distinct from the traditional Bayesian surprise \( S_{\text{Bayesian}} \). While \( S_{\text{Bayesian}} \) measures the shift in model belief using the KL divergence between the prior and posterior distributions, \( S_{\text{Bayesian}_\text{flat}} \) evaluates the shift in model confidence relative to a flat prior baseline. This approach minimizes the influence of potentially biased or overly informative priors, ensuring that the metric remains robust even when prior knowledge is limited or unreliable.  \( S_{\text{Bayesian}_\text{flat}} \) can be expressed as:

\begin{equation}
S_{\text{Bayesian}_\text{flat}} = \text{KL}\left(p(\mathcal{M}_{\text{posterior}}) \parallel p(\mathcal{M}_{\text{flat}})\right)
\end{equation}

In the above equation, \( \mathcal{M}_{\text{flat}} \) refers to a flat prior model, representing a minimally informed baseline for the predictive distribution. Unlike a typical prior, \( \mathcal{M}_{\text{flat}} \) assumes minimal knowledge, reflecting a completely uninformative belief about the outcomes. This flat prior serves as a neutral baseline against which the posterior predictive distribution, \( \mathcal{M}_{\text{posterior}} \) is compared. The confidence correction term, \( C \), accounts for the uncertainty in the GP’s predictions, defined as the negative entropy of the GP’s predictive variance:

\begin{equation}
C = -\frac{1}{2} \log(2 \pi e \sigma_{\text{prior}}^2)
\end{equation}


This term ensures that observations in regions where the model is more confident (i.e., with lower predictive uncertainty) receive greater weight. In contrast, if the model is highly uncertain, the term becomes more negative, discounting the contribution of surprise from those regions. Finally, the adjustment term, \( A \), aligns CAS with a flat prior baseline, ensuring that the CAS is robust to shifts in model certainty. It is calculated as:

\begin{equation}
A = -\log \left(1 - \Phi\left(\frac{\mu_{\text{flat}}}{\sigma_{\text{flat}}}\right)\right)
\end{equation}

where \( \Phi \) is the cumulative distribution function (CDF) of the standard normal distribution, anchoring CAS to a naive observer's baseline belief. \( A \) corrects CAS by adjusting for the expected surprise from a flat, uninformed prior, ensuring that CAS remains robust to shifts in model certainty rather than overreacting to prior belief updates. Unlike traditional metrics that focus solely on unexpected outcomes, CAS incorporates dynamic confidence adjustments. This approach prioritizes experiments that promise high informational value relative to the model’s current state, thereby not only highlighting surprising observations but also identifying those with significant potential to enhance our understanding of the search space.

\subsection{Philosophical Differences Among Surprise Metrics}

While all three measures are rooted in the basic notion of surprise measurement, understanding their philosophical differences is essential before evaluating their potential inclusion as data acquisition strategies in a sequential learning framework and selecting the most effective one under resource constraints. To illustrate on this, consider the task of discovering high-performance alloys, where a GP model predicts material hardness (\( y \)) based on composition and microstructure (\( \mathbf{x} \)). When a new observation \(D_{obs}=\{\mathbf{x_{\text{obs}}}, y_{\text{obs}}\}\) significantly deviates from the GP's prior prediction, each surprise metric interprets its importance differently. Shannon Surprise focuses on the rarity of the observation under the current predictive distribution. For instance, if the predicted hardness for an alloy is 200 MPa but the observed hardness is 400 MPa, Shannon Surprise flags this as highly unexpected. While this highlights anomalies, Shannon Surprise does not consider whether the observation contributes meaningfully to refining the model. Consequently, it may over-prioritize anomalies caused by noise or experimental errors, leading to inefficient resource allocation. Bayesian Surprise, in contrast, measures how much this new observation updates the model's belief, using the KL divergence between prior and posterior distributions. In the same example, if incorporating \(D_{obs}\) significantly shifts the GP’s understanding of the relationship between alloy composition and hardness, Bayesian Surprise assigns a high value. This ensures that only impactful observations receive priority. However, Bayesian Surprise may overvalue updates in regions with sparse data or high uncertainty, potentially diverting attention from more promising areas. CAS, on the other hand, combines the strengths of Shannon Surprise, which evaluates the unexpectedness of an observation, and Bayesian Surprise, which measures its impact on belief updates, while addressing their limitations. It incorporates confidence adjustments to account for predictive uncertainty, dynamically weighing observations based on how confident the model is in its predictions. Additionally, CAS aligns belief updates with a flat prior baseline, anchoring evaluations to a neutral starting point rather than the model’s evolving prior. This ensures robustness in data-scarce environments and prevents over-prioritization of uncertain regions. By balancing unexpectedness, belief updates, and confidence adjustments, CAS achieves an effective trade-off between exploration and exploitation. Unlike Shannon Surprise, which may misinterpret noise as valuable, or Bayesian Surprise, which can overemphasize uncertainty, CAS prioritizes observations that are both surprising and impactful. For example, if the aforementioned observation significantly deviates from predicted properties while challenging foundational assumptions about material behavior, CAS ensures its importance is accurately captured, guiding exploration toward impactful discoveries while avoiding wasted effort on irrelevant anomalies.

\subsection{Proposed Framework: CA-SMART}

Our proposed  active learning framework (CA-SMART) is specifically designed to optimize information gain while minimizing the number of experimental trials, making it ideal for resource-constrained scenarios. At its core, CA-SMART implements a novel acquisition policy based on the CAS metric within a BO framework. This policy dynamically balances exploration and exploitation by prioritizing data points that maximize model refinement potential while accounting for predictive uncertainty and belief updates. Through this carefully designed CAS-driven acquisition strategy, CA-SMART ensures efficient resource utilization and impactful learning. The detailed design and working of CA-SMART are described in the following steps:

\begin{enumerate} [I]
 
    \item \textbf{Initialize the GP Model:} The first step involves creating an initial dataset to train the GP model. To ensure a well-distributed and representative coverage of the design space, this framework utilizes Sobol sampling, a quasi-random sequence known for its low-discrepancy properties \cite{sobol1967distribution}. Unlike purely random sampling, Sobol sequences efficiently fill the input space, even in higher dimensions, ensuring that the initial dataset is evenly distributed. This mitigates the risk of bias in the initial model predictions and improves the reliability of the sequential learning process. The Sobol sequence is used to generate an initial set of candidate points, denoted as \( \mathbf{S} \), which is defined as:
    
    \begin{equation}
    \mathbf{S} = \text{Sobol}(\mathbf{X_{\text{bounds}}}, n_{\text{candidates}})
    \label{eq:sobol_sampling}
    \end{equation}
    
    where \( \mathbf{X_{\text{bounds}}} \) specifies the bounds of the design space, and \( n_{\text{candidates}} \) is the number of candidate points generated. The size of the initial dataset, denoted as \( (\mathbf{X_{\text{init}}}, \mathbf{y_{\text{init}}}) \), depends on the complexity and dimensionality of the design space. For relatively simple or low-dimensional problems, a smaller initial sample size (e.g., 10–15 samples) may suffice. For higher-dimensional spaces, the number of initial samples can be scaled up proportionally to capture more variation and avoid sparse coverage. Once the Sobol sequence is generated, the initial dataset is formed by sampling points from \( \mathbf{S} \) and observing their corresponding outputs. The GP model is then initialized using this dataset to estimate the mean and covariance of the underlying function. After initializing the GP model, the first candidate point \( \mathbf{x_{\text{next}}} \) is randomly selected using the Sobol sampling and added to the list of sampled points. The initial sampled dataset is finally set as \( (\mathbf{X_{\text{sample}}}, \mathbf{y_{\text{sample}}}) \gets \mathbf{(X_{\text{init}}}, \mathbf{y_{\text{init}}}) \).


    \item \textbf{Surprise Evaluation at $\mathbf{x_{\text{next}}}$:} At each iteration, the GP model is updated with the current dataset, $(\mathbf{X_{\text{sample}}}, \mathbf{y_{\text{sample}}})$, incorporating the latest observations to refine its understanding of the design space. The CAS value for the candidate point $\mathbf{x_{\text{next}}}$, denoted as $\text{CAS}_{\mathbf{x_{\text{next}}}}$, is computed based on the GP model’s posterior predictive distribution. This metric quantifies the degree to which the observed response at $\mathbf{x_{\text{next}}}$ deviates from the model's predicted response. A critical component of this evaluation is the threshold, $K_{\text{CAS}}$, which determines whether the observed response at $\mathbf{x_{\text{next}}}$ is flagged as a surprise. For this framework, $K_{\text{CAS}}$ is set based on a 95\% credible interval of the GP model's predictions, although alternative confidence levels can be used depending on the application. Specifically, $K_{\text{CAS}}$ establishes the allowable range of deviation before an observation is considered surprising. The threshold is computed using the posterior mean $\mu_{\mathbf{x_{\text{next}}}}$ and standard deviation $\sigma_{\mathbf{x_{\text{next}}}}$ of the GP model at the candidate point. If the observed response at $\mathbf{x_{\text{next}}}$ falls outside this interval, the CAS value at $\mathbf{x_{\text{next}}}$ is deemed a surprise. The choice of $K_{\text{CAS}}$ affects the sensitivity of the framework to surprises. A lower $K_{\text{CAS}}$ makes the framework more responsive to deviations, identifying more observations as surprises. Conversely, a higher $K_{\text{CAS}}$ reduces sensitivity, flagging only significant deviations as surprises.

    \item \textbf{Decision Rule – Exploitation or Exploration:} In this step, the active learning framework determines whether to exploit or explore based on the CAS value at the candidate point \(\mathbf{x_{\text{next}}}\), denoted as \(\text{CAS}_{\mathbf{x_{\text{next}}}}\). If \(\text{CAS}_{\mathbf{x_{\text{next}}}}\) exceeds the threshold \(K_{\text{CAS}}\), the observation is flagged as a potential surprise, triggering an exploitation step. Otherwise, the algorithm proceeds with exploration.

    During exploitation, the algorithm investigates the nature of the surprise by drawing a new observation in close proximity to \(\mathbf{x_{\text{next}}}\). This step ensures that the surprise is not due to random noise or data corruption but reflects an underlying discrepancy in the response surface. The perturbed candidate point \(\mathbf{x_{\text{perturbed}}}\) is generated as:

    \begin{equation}
    \mathbf{x_{\text{perturbed}}} = \mathbf{x_{\text{next}}} + \epsilon, \quad \epsilon \sim \mathcal{N}(0, \sigma_{\text{perturb}}^2 \mathbf{I})
    \label{eq:perturbation}
    \end{equation}

    where \(\sigma_{\text{perturb}}\) defines the scale of the local neighborhood, and \(\mathbf{I}\) is the identity matrix. If \(\text{CAS}_{\mathbf{x_{\text{perturbed}}}}\) also exceeds \(K_{\text{CAS}}\), the initial observation is confirmed as a true surprise. Both \( (\mathbf{x_{\text{next}}}, y_{\text{next}}) \) and \( (\mathbf{x_{\text{perturbed}}}, y_{\text{perturbed}}) \) are added to the dataset to refine the model. If \(\text{CAS}_{\mathbf{x_{\text{perturbed}}}}\) is below \(K_{\text{CAS}}\), the original observation is likely noise, and only \( (\mathbf{x_{\text{perturbed}}}, y_{\text{perturbed}}) \) is added, switching to exploration mode for the next experiment. This exploitation process commits additional resources to verify surprising observations, ensuring the model avoids misleading conclusions. Though it may initially slow experimentation, it ultimately enhances overall exploration by reducing the impact of incorrect data. Once the verification process is completed, the model will conduct at least one more experiment in the neighborhood (i.e., exploiting) to assess knowledge gain. If the observation still yields surprise, exploitation will continue. Otherwise, the model will transition to exploration, targeting under-sampled regions to maximize information gain.

    On the other hand, during exploration, the algorithm selects a new candidate point to explore uncharted regions of the design space. For this, a candidate set \(\mathbf{S}\) is generated using Sobol sampling as defined in Equation~\ref{eq:sobol_sampling}, ensuring a uniform distribution of points across the design space. Afterward, to select the next experimental location, a maximin strategy is employed (Equation~\ref{eq:maximin}),which maximizes the minimum distance between the candidate point and all previously sampled points \cite{johnson1990minimax}. This prevents clustering and promotes efficient coverage of the design space \cite{joseph2015maximum}.

    \begin{equation}
    \mathbf{x_{\text{next}}} = \arg \max_{\mathbf{x} \in \mathbf{S}} \left( \min_{\mathbf{e} \in \mathbf{X_{\text{sample}}}} \text{BallTree}(\mathbf{x}, \mathbf{e}) \right)
    \label{eq:maximin}
    \end{equation}

    Here, \(\mathbf{X_{\text{sample}}}\) denotes the set of previously observed points, \(\mathbf{e}\) represents a single point from this set, and the BallTree algorithm is used for efficient nearest-neighbor computations \cite{fu2020framework}. BallTree organizes points in a hierarchical structure, allowing fast distance queries and reducing computational overhead, particularly in high-dimensional spaces. By alternating between exploitation to confirm surprises and exploration to investigate uncharted areas, this framework balances local refinement and global exploration, enabling efficient approximation of the response surface. The sequential process continues until the experimental budget is exhausted.

    \item \textbf{Model Update:}
    After each selected experiment, the GP model is updated with the expanded dataset $(\mathbf{X_{\text{sample}}}, \mathbf{y_{\text{sample}}})$. This allows the model to refine its predictive mean and uncertainty estimates based on the latest observations.

    \item \textbf{Stopping Criteria:}
    The process is repeated iteratively until a specified stopping criterion is met, such as reaching the maximum number of iterations, achieving a predefined level of model accuracy, or convergence in the GP model predictions.

\end{enumerate}

\subsection{Function Approximation using the CA-SMART Framework}

\begin{figure*}[!htb]
    \centering
    \includegraphics[width=\linewidth]{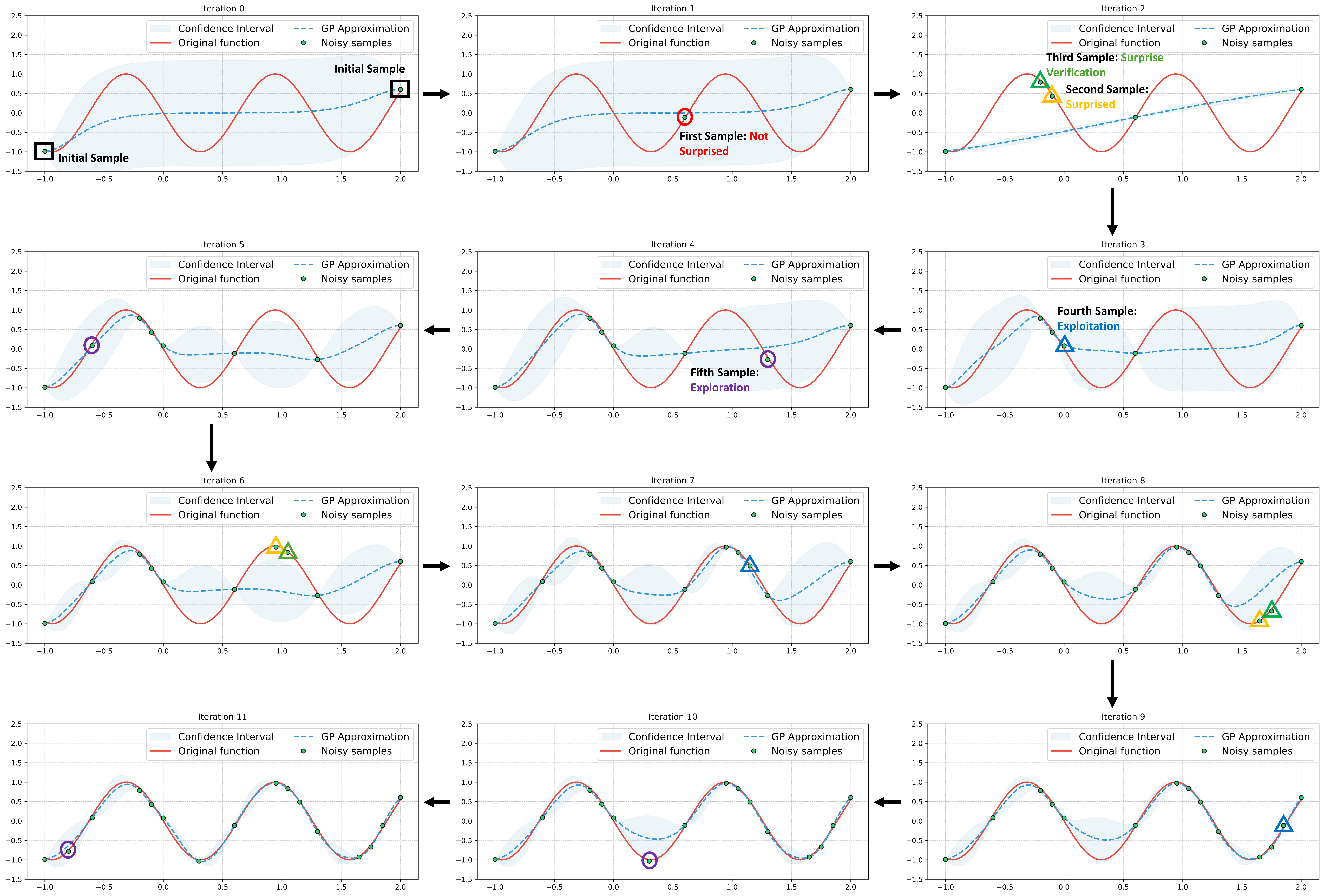}
    \caption{Sequential approximation of the 1D function using the CA-SMART framework. The figure illustrates the progression from initial samples (Iteration 0) to the final approximation (Iteration 11), highlighting the model's dynamic switching between exploration and exploitation driven by the CAS metric. Exploration points (yellow and purple triangles) and exploitation samples (blue and green triangles) refine the model incrementally.}
    \label{fig:model_framework}
\end{figure*}

To illustrate the effectiveness of our proposed framework, we begin by applying the CA-SMART framework to a simple 1D function. This allows us to visually demonstrate the framework's working and provide a clear explanation of its mechanisms. While we use a 1D function for simplicity, the framework is fully capable of handling high-dimensional problems, which will be explored in detail in the next section. The target function is defined as follows:

\begin{equation}
f(x) = -\sin(5x) - 0.75x + 0.75x + \epsilon
\end{equation}

where \( \epsilon \) is a Gaussian noise with standard deviation \( \sigma_{\text{noise}} = 0.05 \). The function combines sinusoidal behavior with a linear component, creating non-trivial patterns for the GP model to approximate. The initial sample points are chosen as \( \mathbf{x} = \{-1.0, 2.0\} \). The sequential learning process, illustrated in Figure \ref{fig:model_framework}, begins with two initial samples (Iteration 0). At each iteration, the CA-SMART framework adaptively switches between exploration and exploitation based on the CAS metric, as described in the previous section. The iterative steps followed by our proposed framework in approximating this function are provided below:

\textbf{Iteration 0:} The process starts with two initial sample points located at \( \mathbf{x} = -1.0 \) and \( \mathbf{x} = 2.0 \). The GP model is initialized using these samples, and the predictive mean, confidence interval, and approximation are shown in Figure \ref{fig:model_framework}.

\textbf{Iteration 1:} A new candidate point is selected randomly through the Sobol sampling expressed in Equation~\ref{eq:sobol_sampling}. The CAS value at this point does not exceed the predefined threshold, indicating that the model is not surprised. Consequently, this sample is added to the dataset, and the GP model is updated.

\textbf{Iteration 2: Exploration and Surprise Detection:} Following an exploration criterion based on selecting the most distant point from previously sampled locations, the second candidate point is identified in a region where the model's uncertainty is high. This time, the CAS value exceeds the threshold, indicating a surprise (highlighted by the yellow triangle). To confirm this observation of surprise, a perturbed sample \( x_{\text{perturbed}} \) is generated near the surprising point using Equation~\ref{eq:perturbation}. The perturbed sample (third sample) also yields a surprise highlighted by the green triangle, verifying the initial surprise observation. Both these samples are added to the dataset.

\textbf{Iteration 3: Exploitation near the Surprising Region:} The GP model is updated with the new samples from Iteration 2. To refine the model's knowledge in the region of interest, the framework enters an exploitation phase. A new sample is selected near the surprising points (blue triangle), further improving the local approximation. The model will undergo exploitation until the surprise value falls below the threshold.

\textbf{Iteration 4: Exploration Continues:} The model continues exploration, selecting a new candidate point in an unexplored region. This time, the CAS value is below the threshold, indicating no surprise. The sample is added to the dataset, and the GP model is updated.

\textbf{Iteration 5-10: Subsequent Iterations:} Guided by the CAS metric, the framework dynamically switches between exploration and exploitation. Surprising points trigger local refinements through exploitation, while non-surprising points encourage further exploration of uncertain regions. As the iterations progress, the model incrementally improves its approximation of the target function.

\textbf{Iteration 11: Final Iteration:} The sequential learning process terminates after 11 iterations. By this point, the GP model has achieved a high-quality approximation of the target function. The confidence intervals have narrowed significantly, and the GP’s predictive mean closely aligns with the true function. 

Figure \ref{fig:model_framework} highlights the effectiveness of the CA-SMART framework in balancing exploration and exploitation. Exploration, indicated by yellow and purple triangles, is guided by high uncertainty, enabling the model to sample regions where confidence is low. Conversely, exploitation, represented by blue and green triangles, is driven by observations of surprise, refining the model in areas with significant deviations between predictions and observations. By dynamically switching between these two strategies, the framework leverages the CAS metric to prioritize sampling in the most informative regions. This iterative process allows the CA-SMART framework to efficiently approximate the unknown function while minimizing the number of samples, showcasing the strength of the CAS-driven active learning approach in resource-constrained scenarios.




\section{Results and Discussion}

\subsection{Performance Metrics}

In order to evaluate the performance of our proposed CA-SMART approach and compare it with other baseline methods, we employ two primary performance metrics: Root Mean Square Error (RMSE) and Continuous Ranked Probability Score (CRPS). These metrics are chosen for their ability to capture the accuracy and reliability of the predictions across iterations. RMSE is a commonly used metric for evaluating the accuracy of point estimates. It is defined as the square root of the mean of the squared differences between predicted values and the actual target values. Given a set of predictions \( \hat{y}_i \) and actual observations \( y_i \) for \( i = 1, \dots, N \), RMSE is calculated as follows:

\begin{equation}
    \text{RMSE} = \sqrt{\frac{1}{N} \sum_{i=1}^{N} (\hat{y}_i - y_i)^2}
\end{equation}

Lower RMSE values indicate that the predicted values are closer to the actual values, suggesting that the model is accurately approximating the true function. In Bayesian Optimization, RMSE provides an indication of the model’s accuracy in approximating the underlying objective function as it explores the search space. 

CRPS provides a more comprehensive assessment of predictive accuracy by evaluating the full probabilistic forecasts rather than just point estimates, as in RMSE. This makes CRPS particularly useful in BO, where uncertainty estimates are available. The CRPS measures the difference between the CDF of the predicted distribution \( F_{\hat{y}_i} \) and the actual observation \( y_i \), and it is defined as:

\begin{equation}
    \text{CRPS} = \frac{1}{N} \sum_{i=1}^{N} \int_{-\infty}^{\infty} \left( F_{\hat{y}_i}(z) - \mathbf{1}_{\{z \geq y_i\}} \right)^2 \, dz
\end{equation}

where \( \mathbf{1}_{\{z \geq y_i\}} \) denotes the indicator function, which is 1 if \( z \geq y_i \) and 0 otherwise; $z$ is a variable over which the CDF is integrated. For practical computation, CRPS can often be derived in closed form when the predictive distribution is normal, simplifying its application in models where Gaussian assumptions hold. CRPS penalizes deviations in the predicted distribution both in terms of location (mean) and spread (variance), providing a more comprehensive assessment of prediction quality than RMSE. A lower CRPS value indicates that the predictive distribution is closer to the observed values in terms of both accuracy and reliability. This metric is especially valuable in Bayesian frameworks, where the uncertainty of predictions is a critical component.

\subsection{Performance on Synthetic Benchmark Functions}

To evaluate the performance of our proposed CA-SMART approach, we have conducted experiments on two synthetic benchmark functions: the Six-Hump Camelback function \cite{kleijnen2012expected} and the Griewank \cite{locatelli2003note} function. These benchmark functions are widely utilized due to their complexity and their ability to challenge optimization algorithms, making them suitable for assessing how well an approach can approximate a response surface under limited experimental budgets.  The Six-Hump Camelback function is a two-dimensional function defined over the bounds \( [-3, 3] \) for \( x_1 \) and \( [-2, 2] \) for \( x_2 \). This function is characterized by multiple local minima, presenting challenges for approximation algorithms to capture its intricate structure effectively. Similarly, the Griewank function is a five-dimensional function with each dimension bounded by \( [-600, 600] \). It features a large number of regularly spaced local minima, testing an algorithm's ability to balance exploration and exploitation during function approximation.

In our experiments, the goal is to approximate these functions as accurately as possible within a constrained budget of sequential experiments. For each function, we have employed a GP model as the surrogate which is configured with a composite kernel comprising a Rational Quadratic kernel multiplied by a Constant kernel. This kernel choice allows the GP to capture both smooth variations and abrupt changes in the underlying function. The kernel hyperparameters are optimized during model fitting using the marginal likelihood of the GP. For both the synthetic functions, the initial samples are generated using the Sobol sequence using Equation~\ref{eq:sobol_sampling}. For the Griewank function, 10 initial samples are used, while for the Six-Hump Camelback function, 5 initial samples are employed. 

After initializing the GP model, the sequential samples are selected iteratively based on the CAS metric. At each iteration, the CA-SMART framework adaptively selects candidate points for exploration or exploitation, following the process detailed in Section 3.5. Candidate points are generated using Sobol sampling, and decisions are guided by the CAS metric, leveraging exploitation for surprising observations and exploration to ensure broad design space coverage. For both these synthetic functions, this sequential process continues for a maximum of 60 iterations, with the GP model being updated at each step using the expanded dataset. The performance of CA-SMART is compared against two other surprise-based methods (Shannon surprise and Bayesian surprise) and four traditional BO acquisition functions (UCB, PI, MV, and EI). For all these approaches, RMSE and CRPS are used as perforamnce metrics. To ensure statistical reliability, we have conducted 30 independent runs with a maximum of 60 iterations for each approach for both these functions.

\begin{figure}[!htb]
    \centering
    \includegraphics[width=\linewidth]{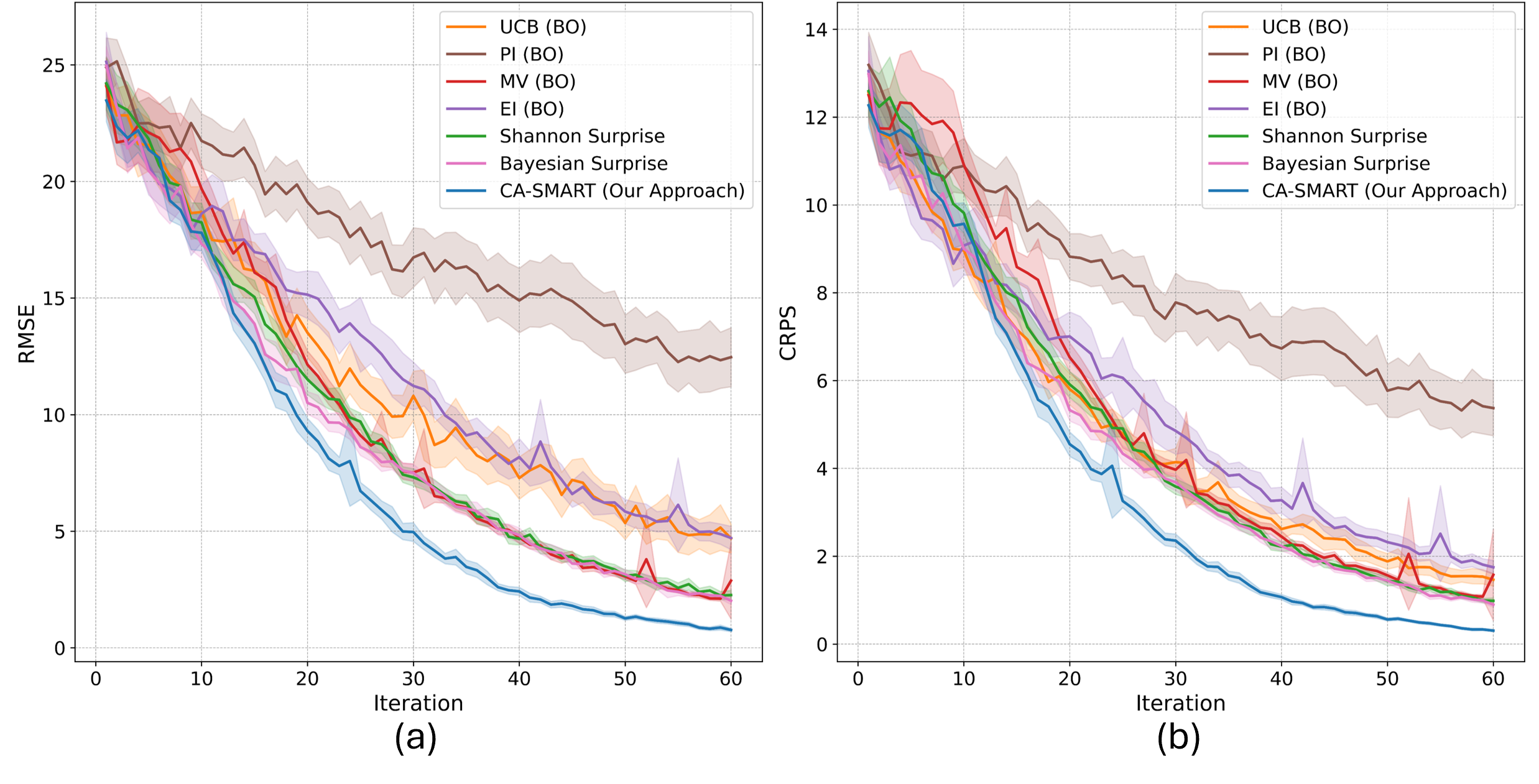}
    \caption{Performance on the Six-Hump Camelback function: (a) RMSE across iterations, (b) CRPS across iterations. Shaded areas represent 95\% confidence intervals.}
    \label{fig:hump_results}
\end{figure}

\begin{figure}[!htb]
    \centering
    \includegraphics[width=\linewidth]{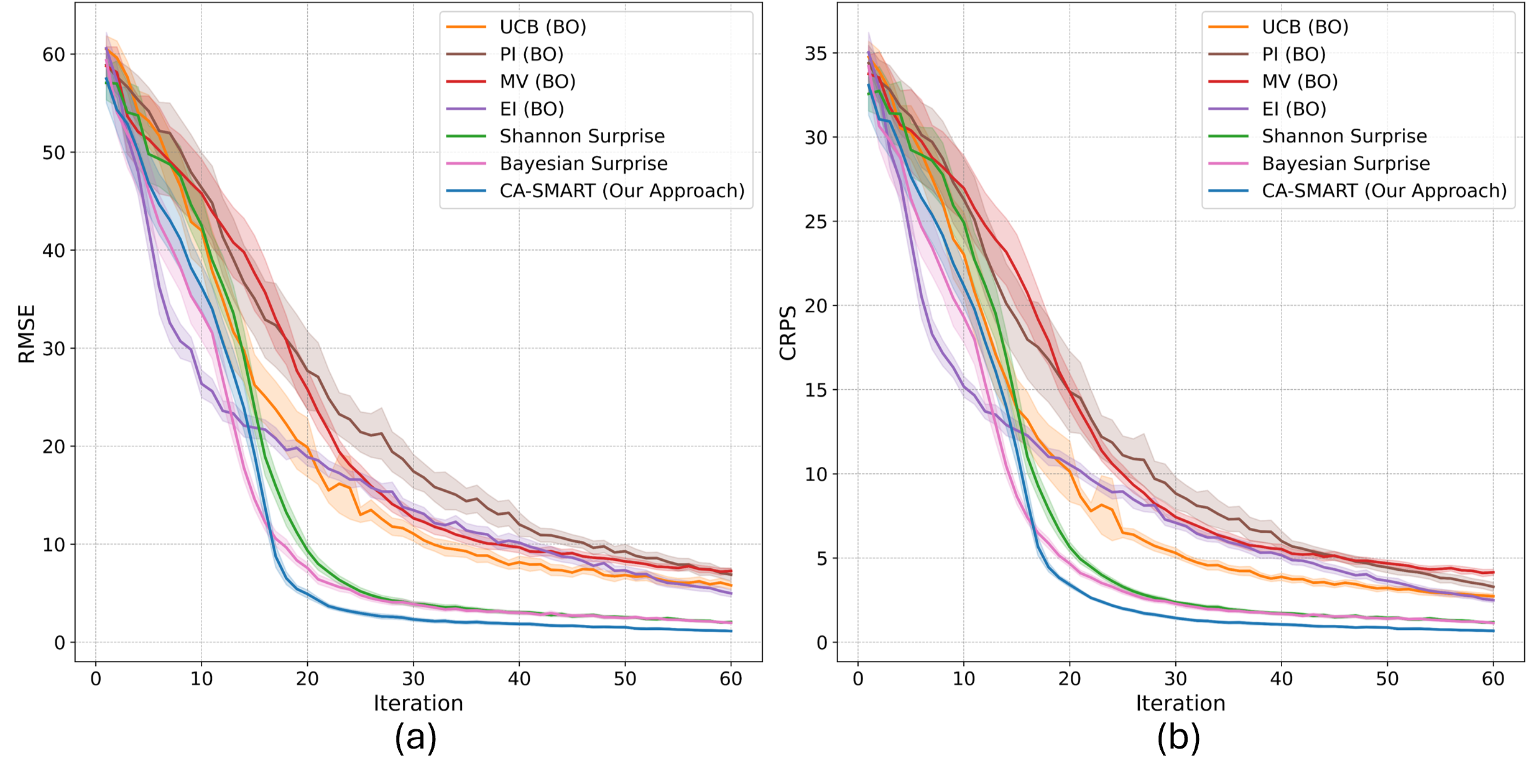}
    \caption{Performance on the Griewank function: (a) RMSE across iterations, (b) CRPS across iterations. Shaded areas represent 95\% confidence intervals.}
    \label{fig:griewank_results}
\end{figure}

Figures~\ref{fig:hump_results} and~\ref{fig:griewank_results} showcase the prediction performance of our proposed CA-SMART framework, along with two surprise-based approaches—Shannon surprise and Bayesian surprise—and four traditional BO methods: UCB, PI, MV, and EI. The shaded regions in all plots represent the 95\% confidence intervals, providing insight into the variability of performance across runs.

Figures~\ref{fig:hump_results}a and ~\ref{fig:hump_results}b display the RMSE and CRPS, respectively, for all approaches over 60 sequential iterations for the Six-Hump Camelback function. This function is particularly challenging due to its multiple local minima, requiring optimization algorithms to balance exploration and exploitation effectively. From the RMSE plot, it is evident that CA-SMART achieves a significantly faster reduction in RMSE compared to other approaches, particularly in the early iterations. Within the first 10 iterations, CA-SMART demonstrates a steep decline in RMSE, indicating its ability to approximate the function efficiently with minimal experimental budget. In contrast, the Shannon Surprise and Bayesian Surprise approaches exhibit similar trends but converge more slowly than CA-SMART. Among the traditional BO methods, MV performs better, with PI performing worst. However, the performance of MV still lags behind the surprise-based methods. The UCB and EI approaches show relatively poorer performance, likely due to their tendencies to favor exploration and exploitation, respectively, without dynamically adapting to the underlying search space. The CRPS results (Figure~\ref{fig:hump_results}b) further reinforce the superior performance of CA-SMART. The CRPS values for CA-SMART decline steadily, demonstrating both accurate predictions and reduced uncertainty over the sequential iterations. Shannon Surprise and Bayesian Surprise follow similar trajectories but remain slightly above CA-SMART. Traditional BO methods, particularly PI and EI, show slower convergence and higher variability, as indicated by their broader confidence intervals. This underscores the limitations of these approaches in effectively handling exploration-exploitation trade-offs for functions with multiple local minima.

Figure~\ref{fig:griewank_results} presents the RMSE (Figure~\ref{fig:griewank_results}a) and CRPS (Figure~\ref{fig:griewank_results}b) results for the Griewank function in a 5D search space. The Griewank function poses a significant challenge due to its large number of regularly spaced local minima, which makes it difficult for optimization strategies to approximate the function accurately without excessive exploration. H After the first 20 iterations, CA-SMART achieves a sharp reduction in RMSE, reflecting its ability to identify informative points. Shannon Surprise and Bayesian Surprise follow closely but exhibit slightly higher RMSE values. The traditional BO methods, including UCB, PI, MV, and EI, converge more slowly and show a plateau in later iterations, indicating suboptimal exploration of the search space. Compared to the suprise-based approaches, they all demonstrate higher variability, suggesting that they may not be efficiently balancing the exploration-exploitation trade-off. The CRPS plot for the Griewank function (Figure~\ref{fig:griewank_results}b) shows a similar trend. CA-SMART achieves the lowest CRPS values with rapid convergence, indicating superior predictive accuracy and uncertainty quantification. Shannon Surprise and Bayesian Surprise again perform well but remain slightly behind CA-SMART. The traditional BO methods exhibit slower convergence, with PI and MV showing the poorest performance due to their biases toward exploitation and exploration, respectively.

Across both benchmark functions, the results consistently highlight the advantages of CA-SMART in terms of both RMSE and CRPS metrics. The rapid convergence of CA-SMART, especially in the early iterations, demonstrates its efficiency in approximating complex functions under limited sequential evaluations. The incorporation of the CAS metric enables CA-SMART to adaptively balance exploration and exploitation by reacting to surprising and informative observations. In contrast, traditional BO methods, while effective to some extent, fail to achieve similar levels of performance due to their static acquisition strategies. These results validate the robustness and efficiency of CA-SMART in approximating both low-dimensional and high-dimensional benchmark functions, outperforming both surprise-based and traditional BO approaches in terms of accuracy, convergence speed, and uncertainty quantification.

\begin{figure}[!htb]
    \centering
    \includegraphics[width=\linewidth]{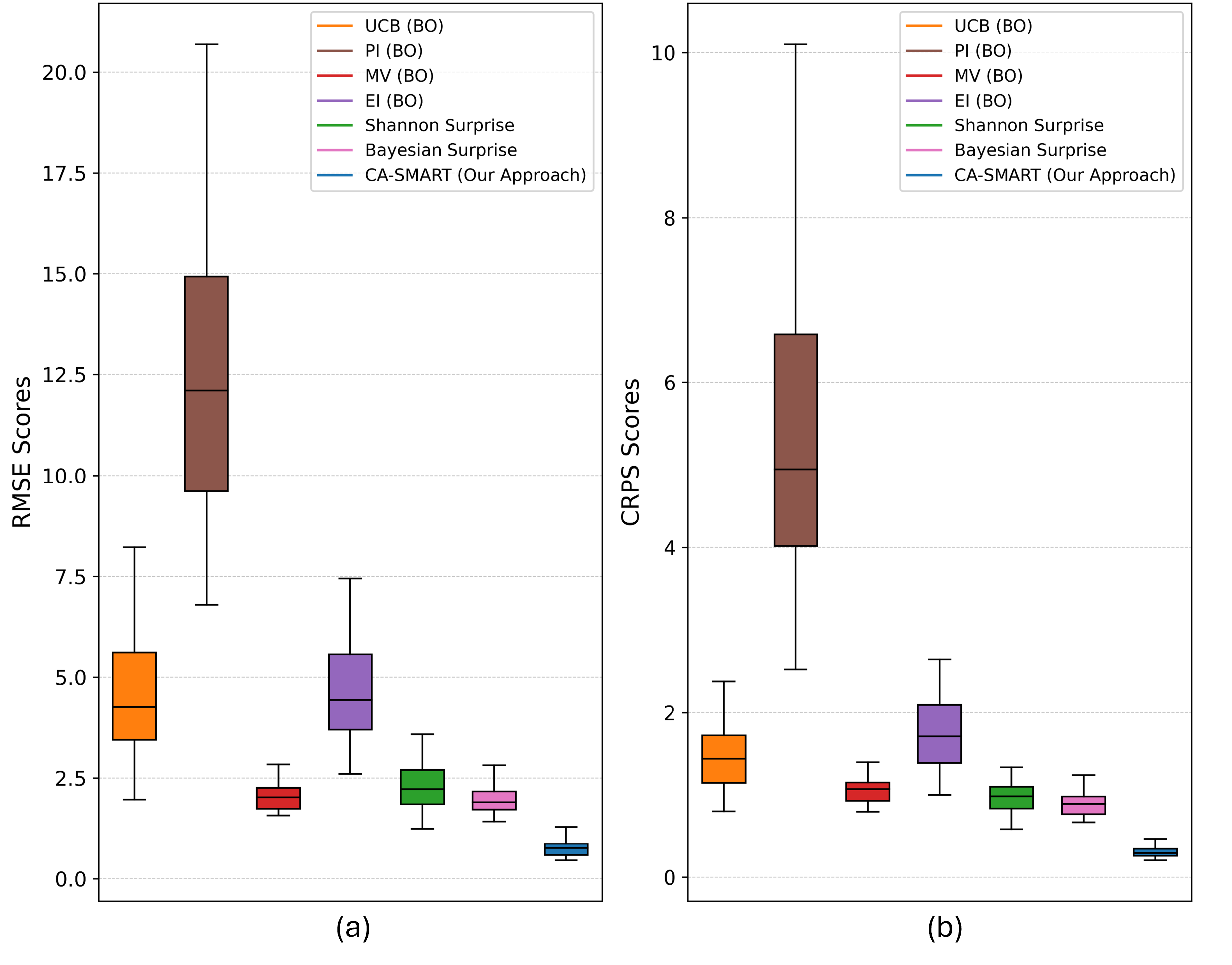}
    \caption{Final performance comparison on the Six-Hump Camelback function at the 60th iteration: (a) RMSE scores across 30 runs, (b) CRPS scores across 30 runs.}
    \label{fig:hump_boxplots}
\end{figure}

\begin{figure}[!htb]
    \centering
    \includegraphics[width=\linewidth]{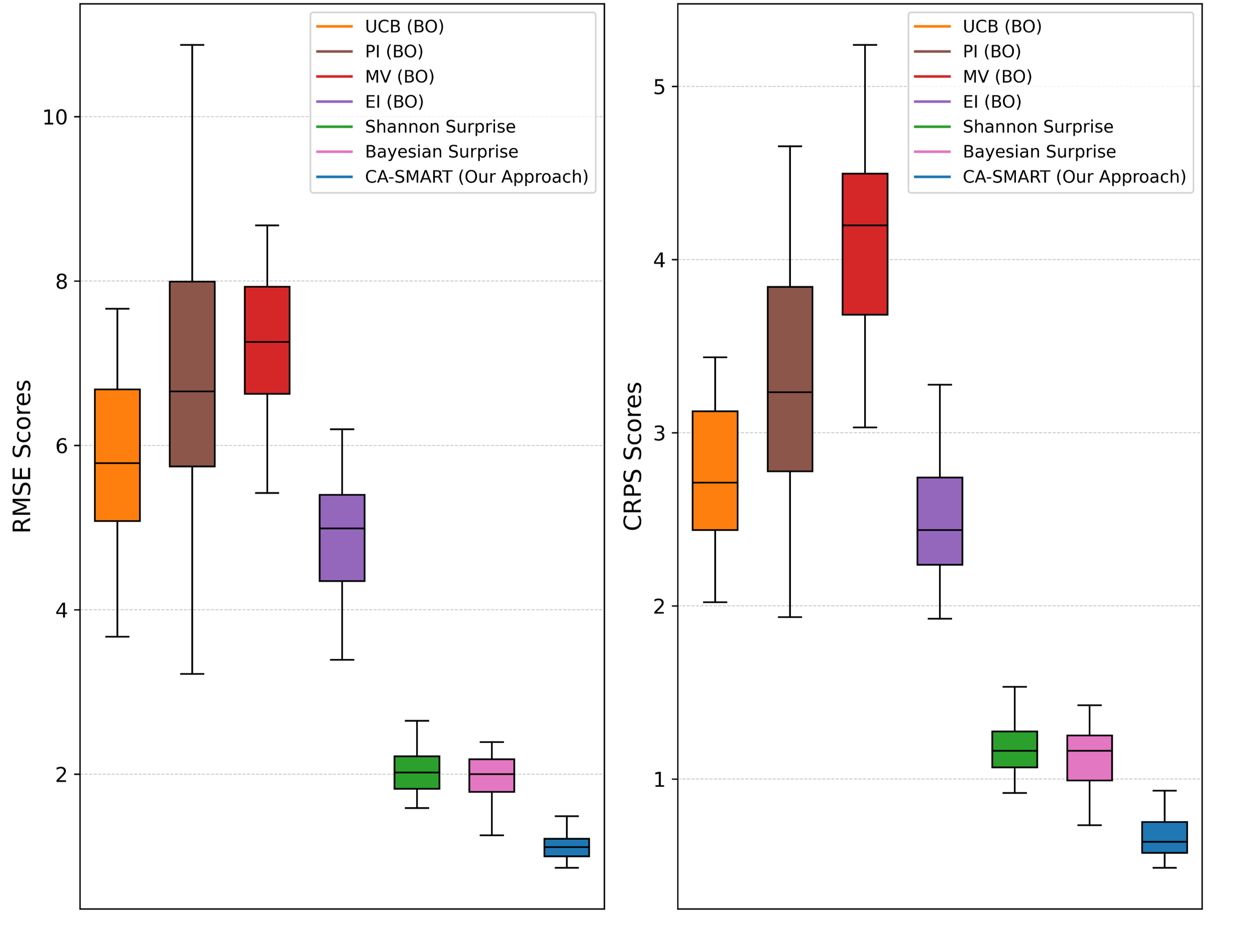}
    \caption{Final performance comparison on the Griewank function at the 60th iteration: (a) RMSE scores across 30 runs, (b) CRPS scores across 30 runs.}
    \label{fig:griewank_boxplots}
\end{figure}

To further analyze the performance of each approach, we present the final RMSE and CRPS distributions at the end of the 60th iteration for both the Six-Hump Camelback and Griewank functions. Figures~\ref{fig:hump_boxplots} and~\ref{fig:griewank_boxplots} display boxplots summarizing the RMSE and CRPS scores across 30 independent runs, providing deeper insights into the consistency and robustness of the tested approaches.

In Figures~\ref{fig:hump_boxplots}a and ~\ref{fig:hump_boxplots}b, which corresponds to the Six-Hump Camelback function, CA-SMART consistently achieves the lowest median RMSE and CRPS scores among all approaches. The boxplots demonstrate that the variability in CA-SMART’s results is also minimal, as reflected by the narrower interquartile range (IQR) and shorter whiskers compared to other methods. This indicates that CA-SMART not only approximates the function more accurately but also does so with a high degree of stability across repeated runs. In contrast, traditional BO approaches, such as EI, PI, UCB, and MV, exhibit higher median scores with wider variability. Notably, the EI and PI acquisition functions struggle to balance exploration and exploitation, leading to less consistent performance. Shannon Surprise and Bayesian Surprise approaches show competitive performance but are still outperformed by CA-SMART, which benefits from its adaptive decision-making mechanism.

Figures~\ref{fig:griewank_boxplots}a and ~\ref{fig:griewank_boxplots}b present the RMSE and CRPS scores for the Griewank function in five dimensions. The results further highlight the superiority of CA-SMART, which again achieves the lowest median RMSE and CRPS values with minimal variability across the 30 runs. For such a complex function with numerous local minima, the ability of CA-SMART to balance exploration and exploitation proves highly advantageous, allowing the model to approximate the function efficiently with fewer sequential iterations. Traditional BO methods, particularly MV and PI, display significant variability, as evidenced by their wider boxplots and higher whiskers. EI and UCB perform moderately well but still lag behind surprise-based approaches, which explicitly prioritize learning from unexpected observations. Shannon Surprise and Bayesian Surprise exhibit better performance compared to traditional BO methods but remain less effective than CA-SMART, reinforcing the value of the confidence correction and adaptive exploration strategy in CA-SMART.

Overall, the results demonstrate that CA-SMART outperforms both surprise-based and BO methods with traditional acquisition functions on complex benchmark functions. Its superior performance is attributed to its enhanced exploration-exploitation balance, which allows it to effectively navigate the search space and avoid local minima.

\subsection{Predicting Fatigue Strength of Steel}

Predicting the fatigue strength of steel is crucial due to its direct impact on the durability and reliability of steel components used in various structural and industrial applications. Fatigue failure accounts for a significant portion of material failures, making accurate predictions essential for improving material design and enhancing safety in real-world applications. Traditional fatigue testing methods are time-consuming and costly, prompting researchers to explore machine learning techniques to predict fatigue strength more efficiently.

The dataset for predicting fatigue strength of steel includes a total of 26 variables. These variables encompass a range of parameters relevant to the material's composition, processing, and structural characteristics. Specifically, the dataset comprises 10 composition parameters, such as elements like carbon (C), silicon (Si), manganese (Mn), and nickel (Ni), which are known to impact the microstructure and strength of steel. Additionally, 12 heat-treatment parameters are included, such as Normalizing Temperature (NT), Carburization Temperature (CT), Diffusion Temperature (DT), and Quenching Media Temperature (QMT), which influence the thermal and mechanical properties of steel. The dataset also includes a rolling parameter, the Reduction Ratio (RR), which affects the steel’s grain structure and overall mechanical properties. Three inclusion parameters, denoted as APID, APIO, and APII, are also present, representing the type and size of inclusions in the steel that can influence its fatigue performance. Finally, the target variable is Fatigue Strength (FS), the primary outcome of interest. 

During the data-cleaning process, any samples with numerical anomalies or missing values were removed, resulting in a refined dataset of 437 samples. To ensure consistency across features and eliminate disparities in magnitude, normalization was applied to all variables, rendering the dataset dimensionless and facilitating a fair comparison among features. Feature selection was performed using a Random Forest Regressor, optimized through GridSearchCV, to identify the most influential features for predicting FS. The optimal hyperparameters were found to be: \texttt{max\_depth} = 10, \texttt{min\_samples\_split} = 2, and \texttt{n\_estimators} = 400. This configuration yielded a Mean Squared Error (MSE) of 596.77 and an $R^2$ score of 0.986, indicating strong predictive capability. To simplify the model and enhance interpretability, we focused on the top 6 most important features. Table~\ref{tab:top_features} lists these features alongside a brief description of each. The selection of these features was guided by their high importance scores, suggesting a significant influence on FS.

\begin{table}[!htb]
    \centering
    \small
    \caption{Top 6 important features for predicting fatigue strength.}
    \label{tab:top_features}
    \begin{tabular}{ll}
        \hline
        \textbf{Feature} & \textbf{Description} \\
        \hline
        NT & Normalizing Temperature \\
        CT & Carburization Temperature \\
        Cr & Chromium Content \\
        QmT & Quenching Media Temperature \\
        DT & Diffusion Temperature \\
        Ct & Carburization Time \\
        \hline
    \end{tabular}
\end{table}

\begin{figure}[!htb]
    \centering
    \includegraphics[width=\columnwidth]{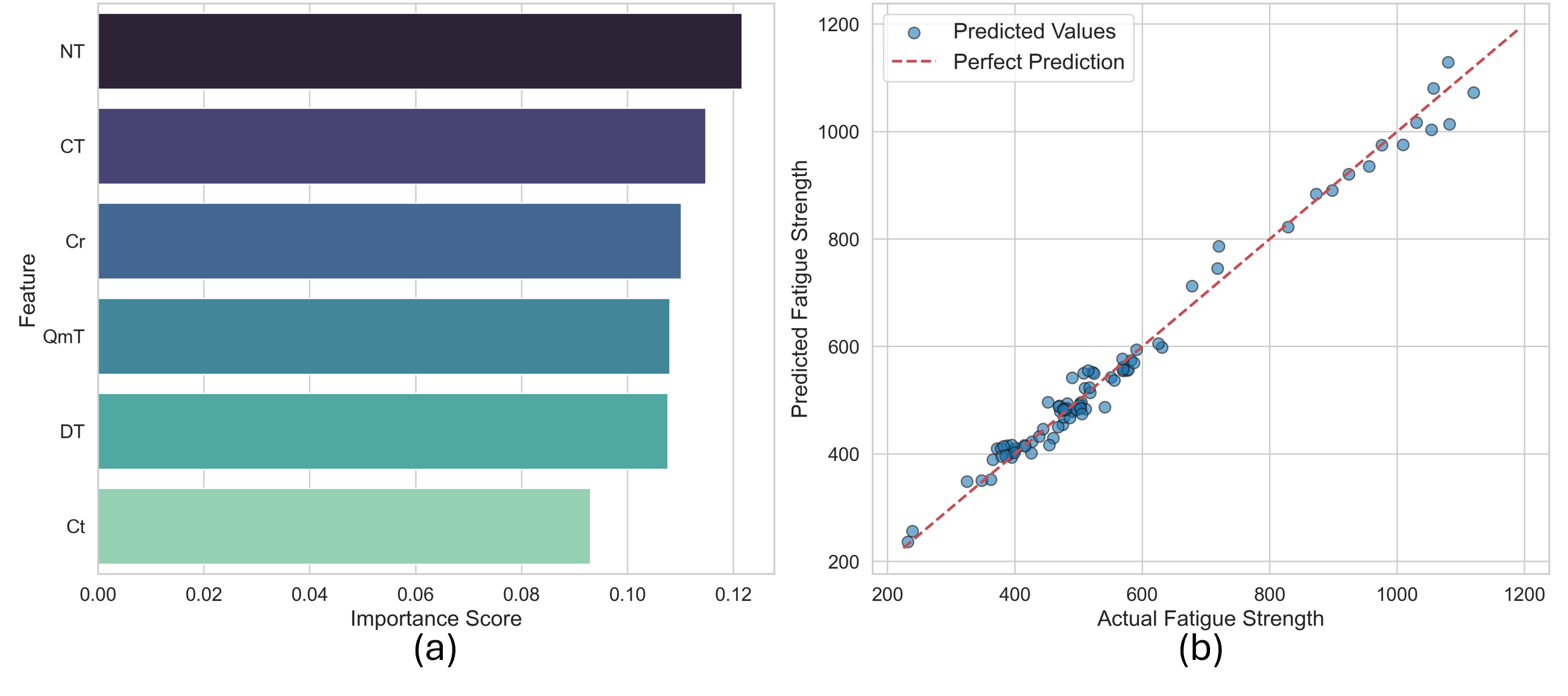}
    \caption{(a) Feature importance scores for predicting fatigue strength. (b) Predicted vs actual fatigue strength values.}
    \label{fig:feature_importance}
\end{figure}

Figure~\ref{fig:feature_importance}a illustrates the importance scores of the top 6 features, while Figure~\ref{fig:feature_importance}b shows the scatter plot of predicted versus actual fatigue strength values. The selected features capture key aspects of composition and heat-treatment parameters, which align with findings from previous studies on steel fatigue strength \cite{liu2023prediction}. These features are crucial as they directly influence the microstructural changes in steel, impacting its fatigue performance.

\subsubsection{Prediction Performance of CA-SMART and Other Approaches}

To evaluate the performance of our proposed CA-SMART approach, we have conducted a comprehensive comparison with two other surprise-based methods, namely Shannon Surprise and Bayesian Surprise, as well as four popular BO acquisition functions: UCB, PI, MV, and EI. For all methods, the GP model is employed as the surrogate model, utilizing the Matern kernel to capture the underlying function's smoothness and complexity effectively. Each approach begins with an initial GP model fitted using 25 samples, followed by sequential updates with 125 additional samples drawn from an available set of 350 candidate samples. The test size is kept constant across all approaches to ensure a fair and unbiased comparison. To account for the inherent randomness in sampling and model predictions, each method is run 30 independent times, minimizing bias and mitigating the impact of stochastic variations in model performance.

\begin{figure}[!htb]
    \centering
    \includegraphics[width=\linewidth]{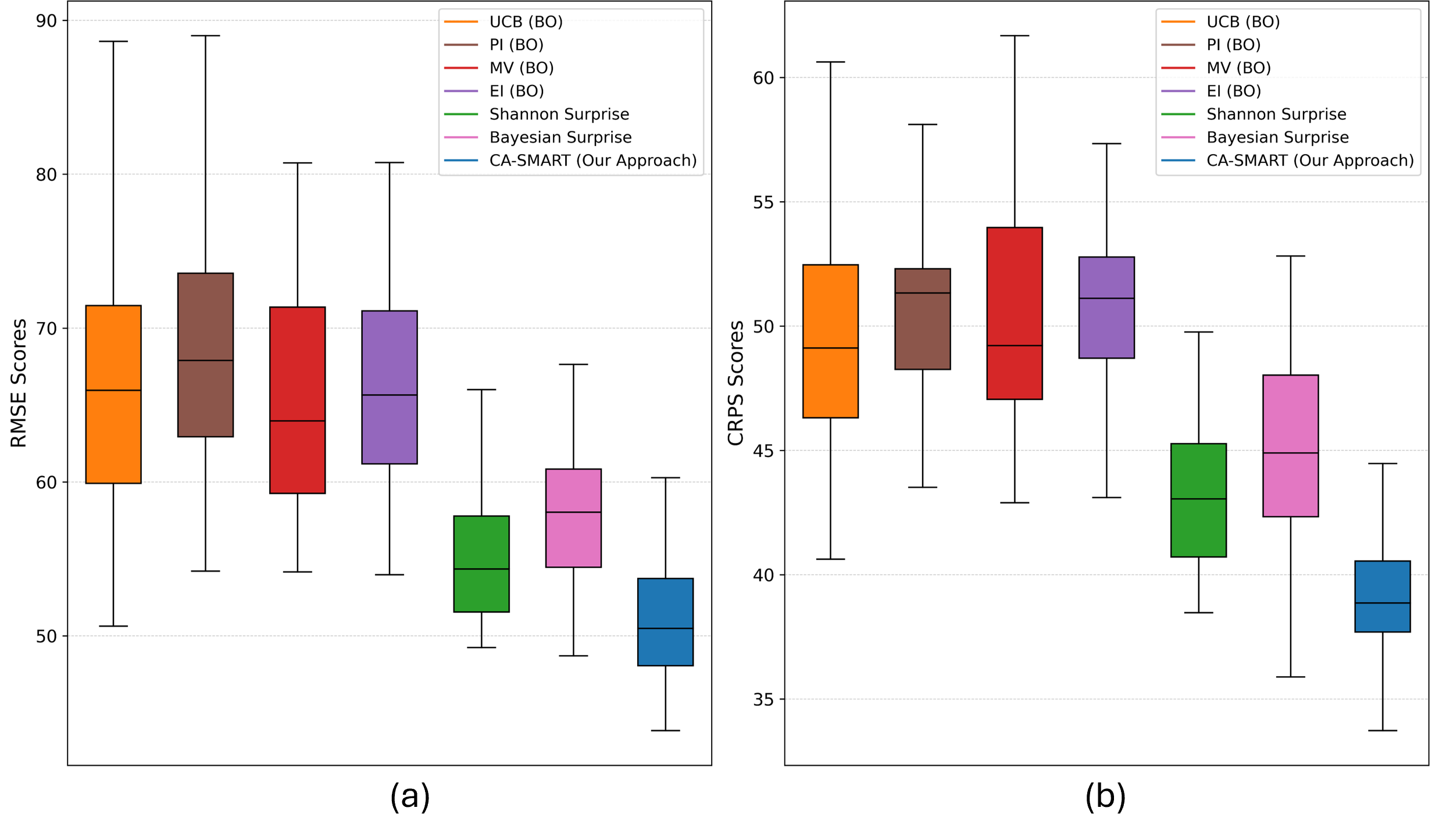}
    \caption{Comparison of model performance on fatigue strength prediction: (a) RMSE scores across 30 runs, (b) CRPS scores across 30 runs.}
    \label{fig:model_comparison_boxplots_fatigue}
\end{figure}

The boxplots in Figure~\ref{fig:model_comparison_boxplots_fatigue} illustrate the distribution of RMSE and CRPS scores across the 30 runs for each method. These visualizations highlight the variability and central tendency of prediction errors, offering insights into the consistency and reliability of each approach. In addition, Table~\ref{tab:mean_ci_fatigue} summarizes the mean and 95\% confidence intervals of RMSE and CRPS for all methods, providing a quantitative assessment of their prediction performance.

\begin{table}[!htb]
    \centering
    \caption{Mean and 95\% confidence intervals of RMSE and CRPS for different approaches.}
    \label{tab:mean_ci_fatigue}
    \resizebox{\columnwidth}{!}{ 
        \begin{tabular}{lcc}
            \hline
            \textbf{Approach} & \textbf{RMSE (Mean ± 95\% CI)} & \textbf{CRPS (Mean ± 95\% CI)} \\
            \hline
            UCB (BO) & 66.08 ± 3.21 & 49.64 ± 1.67 \\
            PI (BO) & 69.65 ± 4.00 & 51.91 ± 2.04 \\
            MV (BO) & 66.08 ± 3.19 & 50.21 ± 1.70 \\
            EI (BO) & 67.27 ± 3.18 & 50.88 ± 1.54 \\
            Shannon Surprise & 55.17 ± 1.53 & 43.38 ± 1.15 \\
            Bayesian Surprise & 58.01 ± 1.81 & 44.80 ± 1.43 \\
            CA-SMART & \textbf{50.76 ± 1.58} & \textbf{39.12 ± 0.95} \\
            \hline
        \end{tabular}
    }
\end{table}

From Figure~\ref{fig:model_comparison_boxplots_fatigue} and Table~\ref{tab:mean_ci_fatigue}, it is evident that CA-SMART consistently achieves superior performance in both RMSE and CRPS metrics compared to other methods. Specifically, CA-SMART yields the lowest RMSE and CRPS mean values with narrow confidence intervals, highlighting both its accuracy and stability. The traditional BO acquisition functions (UCB, PI, MV, EI) exhibit higher RMSE and CRPS scores, suggesting less reliable predictive performance. The Shannon and Bayesian Surprise methods perform better than traditional BO acquisition functions but still lag behind CA-SMART.

The results demonstrate the effectiveness of CA-SMART in accurately predicting fatigue strength, with minimal variability across runs. This superior performance can be attributed to CA-SMART’s improved exploration-exploitation balance, which enables more efficient navigation of the search space and prevents entrapment in local minima. These findings reinforce the robustness and efficiency of CA-SMART for fatigue strength prediction tasks.

\subsubsection{Statistical Analysis of RMSE Scores}

In order to quantitatively assess the reliability and significance of the observed prediction performance, we conduct a comprehensive statistical analysis using the RMSE metric obtained from 30 independent runs for each of the competing sequential learning approaches. The primary objective of these tests is to determine whether the differences in prediction accuracy among the methods are statistically significant, thereby reinforcing the superior performance of the CA-SMART framework.

We first evaluate the normality of the RMSE distributions for each method using the Shapiro--Wilk test \cite{hanusz2014simulation}. Following this, we perform a one-way analysis of variance (ANOVA) to test the null hypothesis that all methods have equal mean RMSE values \cite{das2022analysis}. Upon obtaining a significant ANOVA result, we further employ Tukey’s Honest Significant Difference (HSD) post-hoc test to conduct pairwise comparisons among the methods \cite{nanda2021multiple}. These tests collectively validate the statistical significance of the RMSE differences observed across the 30 runs and support the conclusion that CA-SMART achieves lower prediction errors relative to the other approaches. 

Table~\ref{table:shapiro} summarizes the Shapiro–Wilk normality test for the RMSE scores of the seven sequential learning approaches over 30 independent runs. Notably, while PI (BO) shows a p-value of 0.0121, the RMSE distributions for the remaining methods do not significantly deviate from normality (p $>$ 0.05). Given the robustness of ANOVA to moderate deviations from normality, we proceed with the analysis. A one-way ANOVA (F = 24.8328, p $<$ 0.0001) confirms that the mean RMSE values differ significantly across the methods. To identify which specific pairs of methods differ, we perform Tukey’s HSD post-hoc test. The results of the Tukey test are presented in Table~\ref{table:tukey}. The Tukey HSD results indicate that CA-SMART achieves significantly lower RMSE values compared to all other sequential learning approaches. We can also see that while Shannon Surprise attains the second best performance, yet its RMSE remains statistically higher than that of CA-SMART. These findings demonstrate that the CA-SMART framework consistently reduces prediction error on the fatigue dataset. The significant differences, as determined by both the ANOVA and Tukey HSD tests, support the conclusion that CA-SMART outperforms the competing sequential learning approaches, thereby validating its effectiveness in achieving improved prediction accuracy under resource constraints.

\begin{table}[!htb]
\centering
\small 
\caption{Shapiro–Wilk Normality Test Results for RMSE Scores}
\label{table:shapiro}
\begin{tabular}{lcc}
\hline
\textbf{Method} & \textbf{W-statistic} & \textbf{p-value} \\
\hline
UCB (BO) & 0.9545 & 0.2234 \\
PI (BO) & 0.9064 & 0.0121 \\
MV (BO) & 0.9315 & 0.0537 \\
EI (BO) & 0.9335 & 0.0609 \\
Shannon Surprise & 0.9851 & 0.9389 \\
Bayesian Surprise & 0.9715 & 0.5818 \\
CA-SMART & 0.9767 & 0.7323 \\
\hline
\end{tabular}
\end{table}

\begin{table*}[!htb]
\centering
\small 
\caption{Tukey's HSD Post-Hoc Test Results for RMSE}
\label{table:tukey}
\begin{tabular}{llccccc}
\hline
\textbf{Group1} & \textbf{Group2} & \textbf{Mean Diff.} & \textbf{p-adj} & \textbf{Lower} & \textbf{Upper} & \textbf{Reject} \\
\hline
Bayesian Surprise    & \textbf{CA-SMART}          & -7.5881 & 0.0037 & -13.5485 & -1.6276 & Yes \\
Bayesian Surprise    & EI (BO)           & 9.2641  & 0.0001 & 3.3036   & 15.2245 & Yes \\
Bayesian Surprise    & MV (BO)           & 8.0712  & 0.0015 & 2.1107   & 14.0316 & Yes \\
Bayesian Surprise    & PI (BO)           & 11.6428 & 0.0000 & 5.6824   & 17.6032 & Yes \\
Bayesian Surprise    & Shannon Surprise  & -1.3704 & 0.9933 & -7.3308  & 4.5901  & No  \\
Bayesian Surprise    & UCB (BO)          & 8.0729  & 0.0015 & 2.1124   & 14.0333 & Yes \\
\textbf{CA-SMART}             & EI (BO)           & 16.8521 & 0.0000 & 10.8917  & 22.8126 & Yes \\
\textbf{CA-SMART}             & MV (BO)           & 15.6593 & 0.0000 & 9.6988   & 21.6197 & Yes \\
\textbf{CA-SMART}             & PI (BO)           & 19.2309 & 0.0000 & 13.2704  & 25.1913 & Yes \\
\textbf{CA-SMART}             & Shannon Surprise  & 6.2177  & 0.0347 & 0.2572   & 12.1781 & Yes \\
\textbf{CA-SMART}             & UCB (BO)          & 15.6610 & 0.0000 & 9.7005   & 21.6214 & Yes \\
EI (BO)              & MV (BO)           & -1.1929 & 0.9969 & -7.1533  & 4.7676  & No  \\
EI (BO)              & PI (BO)           & 2.3787  & 0.8978 & -3.5817  & 8.3392  & No  \\
EI (BO)              & Shannon Surprise  & -10.6345& 0.0000 & -16.5949 & -4.6740 & Yes \\
EI (BO)              & UCB (BO)          & -1.1912 & 0.9969 & -7.1516  & 4.7693  & No  \\
MV (BO)              & PI (BO)           & 3.5716  & 0.5693 & -2.3888  & 9.5321  & No  \\
MV (BO)              & Shannon Surprise  & -9.4416 & 0.0001 & -15.4020 & -3.4811 & Yes \\
MV (BO)              & UCB (BO)          & 0.0017  & 1.0000 & -5.9588  & 5.9621  & No  \\
PI (BO)              & Shannon Surprise  & -13.0132& 0.0000 & -18.9736 & -7.0528 & Yes \\
PI (BO)              & UCB (BO)          & -3.5699 & 0.5606 & -9.5304  & 2.3905  & No  \\
Shannon Surprise     & UCB (BO)          & 9.4433  & 0.0001 & 3.4828   & 15.4037 & Yes \\
\hline
\end{tabular}
\end{table*}

Similar to our RMSE analysis, we have applied Shapiro–Wilk normality tests and a one-way ANOVA to the CRPS data across 30 independent runs, confirming that the data meet the assumptions of normality and homogeneity of variances. Table~\ref{table:crps} presents the Tukey HSD post-hoc test results for CRPS. These results further demonstrate that CA-SMART consistently attains the lowest CRPS values across all runs, underscoring its robust predictive reliability and overall superiority compared to the alternative sequential learning approaches.

\begin{table*}[!htb]
\centering
\small 
\caption{Tukey's HSD Post-Hoc Test Results for CRPS}
\label{table:crps}
\begin{tabular}{llccccc}
\hline
\textbf{Group1} & \textbf{Group2} & \textbf{Mean Diff.} & \textbf{p-adj} & \textbf{Lower} & \textbf{Upper} & \textbf{Reject} \\
\hline
Bayesian Surprise & \textbf{CA-SMART} & -5.6848 & 0.0000 & -8.9832 & -2.3864 & Yes \\
Bayesian Surprise & EI (BO) & 6.0745 & 0.0000 & 2.7761 & 9.3729 & Yes \\
Bayesian Surprise & MV (BO) & 5.4080 & 0.0000 & 2.1096 & 8.7064 & Yes \\
Bayesian Surprise & PI (BO) & 7.1044 & 0.0000 & 3.8060 & 10.4027 & Yes \\
Bayesian Surprise & Shannon Surprise & -1.4231 & 0.8582 & -4.7215 & 1.8753 & No \\
Bayesian Surprise & UCB (BO) & 4.8416 & 0.0004 & 1.5432 & 8.1400 & Yes \\
\textbf{CA-SMART} & EI (BO) & 11.7593 & 0.0000 & 8.4609 & 15.0576 & Yes \\
\textbf{CA-SMART} & MV (BO) & 11.0928 & 0.0000 & 7.7944 & 14.3911 & Yes \\
\textbf{CA-SMART} & PI (BO) & 12.7891 & 0.0000 & 9.4908 & 16.0875 & Yes \\
\textbf{CA-SMART} & Shannon Surprise & 4.2617 & 0.0030 & 0.9633 & 7.5601 & Yes \\
\textbf{CA-SMART} & UCB (BO) & 10.5264 & 0.0000 & 7.2280 & 13.8248 & Yes \\
EI (BO) & MV (BO) & -0.6665 & 0.9967 & -3.9649 & 2.6319 & No \\
EI (BO) & PI (BO) & 1.0299 & 0.9673 & -2.2685 & 4.3283 & No \\
EI (BO) & Shannon Surprise & -7.4976 & 0.0000 & -10.7959 & -4.1992 & Yes \\
EI (BO) & UCB (BO) & -1.2329 & 0.9235 & -4.5312 & 2.0655 & No \\
MV (BO) & PI (BO) & 1.6964 & 0.7254 & -1.6020 & 4.9948 & No \\
MV (BO) & Shannon Surprise & -6.8311 & 0.0000 & -10.1294 & -3.5327 & Yes \\
MV (BO) & UCB (BO) & -0.5664 & 0.9987 & -3.8647 & 2.7320 & No \\
PI (BO) & Shannon Surprise & -8.5275 & 0.0000 & -11.8258 & -5.2291 & Yes \\
PI (BO) & UCB (BO) & -2.2627 & 0.3910 & -5.5611 & 1.0356 & No \\
Shannon Surprise & UCB (BO) & 6.2647 & 0.0000 & 2.9663 & 9.5631 & Yes \\
\hline
\end{tabular}
\end{table*}

\subsubsection{Effect of Initial Sample Size on Model Performance}

\begin{figure}[!htb]
    \centering
    \includegraphics[width=\linewidth]{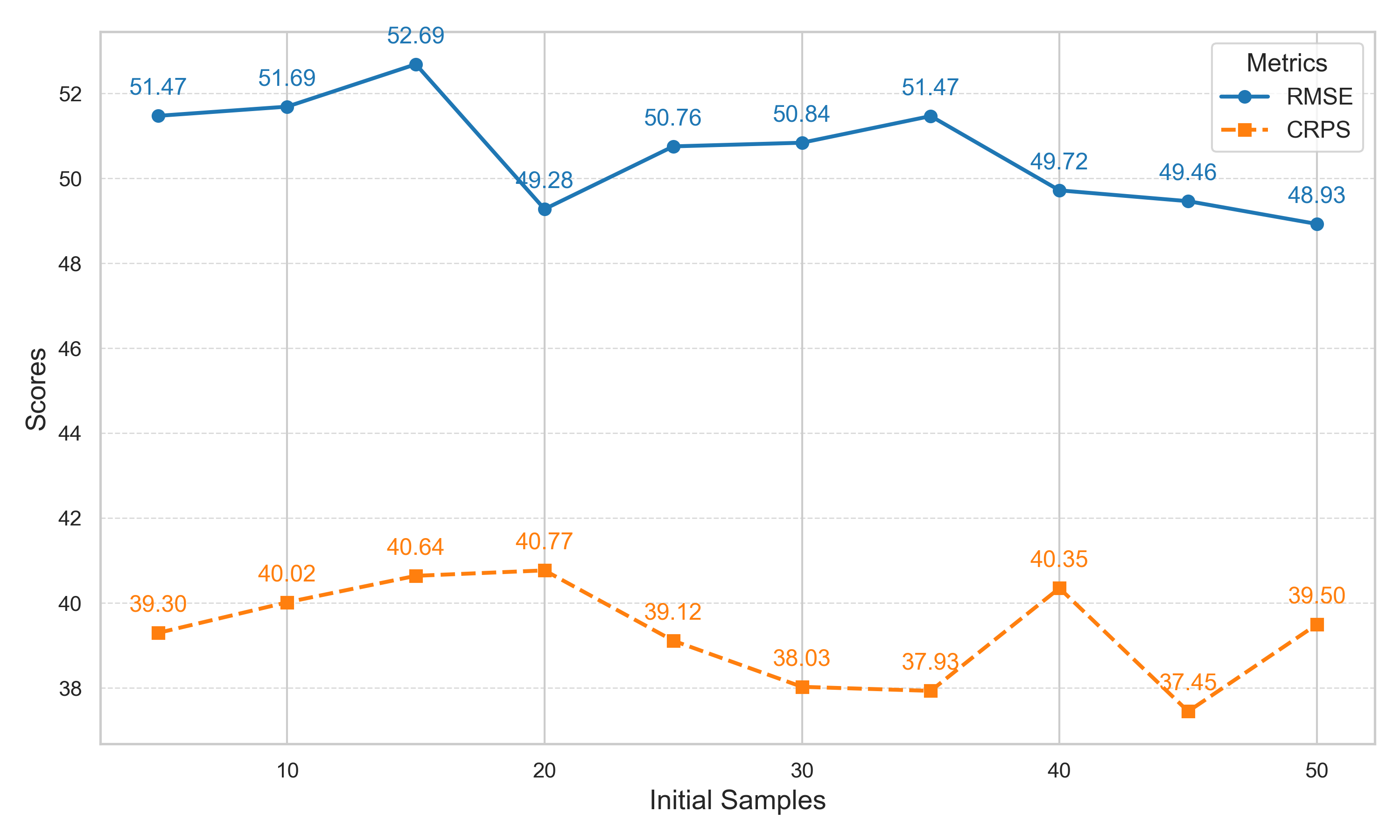}
    \caption{Effect of Initial Sample Size on RMSE and CRPS for CA-SMART}
    \label{fig:initial_sample_effect}
\end{figure}

To investigate the impact of the initial sample size on the performance of our CA-SMART framework, we vary the number of initial samples used to fit the initial GP model keeping the sequential budget fixed to 125 samples. Figure~\ref{fig:initial_sample_effect} presents the RMSE and CRPS scores as a function of the initial sample size. As observed in Figure~\ref{fig:initial_sample_effect}, the number of initial samples has minimal influence on the model’s predictive performance, as indicated by the relatively stable RMSE and CRPS values across different initial sample sizes. Specifically, RMSE scores remain around 50 with slight fluctuations, while CRPS scores stay within a close range, showing only minor variations. This stability demonstrates the robustness of the CA-SMART framework, which is capable of achieving high predictive performance even with a small initial sample size. Consequently, the CA-SMART approach is effective in data-scarce environments, as it does not require a large number of initial samples to start producing reliable predictions.

\begin{figure}[!htb]
    \centering
    \includegraphics[width=\linewidth]{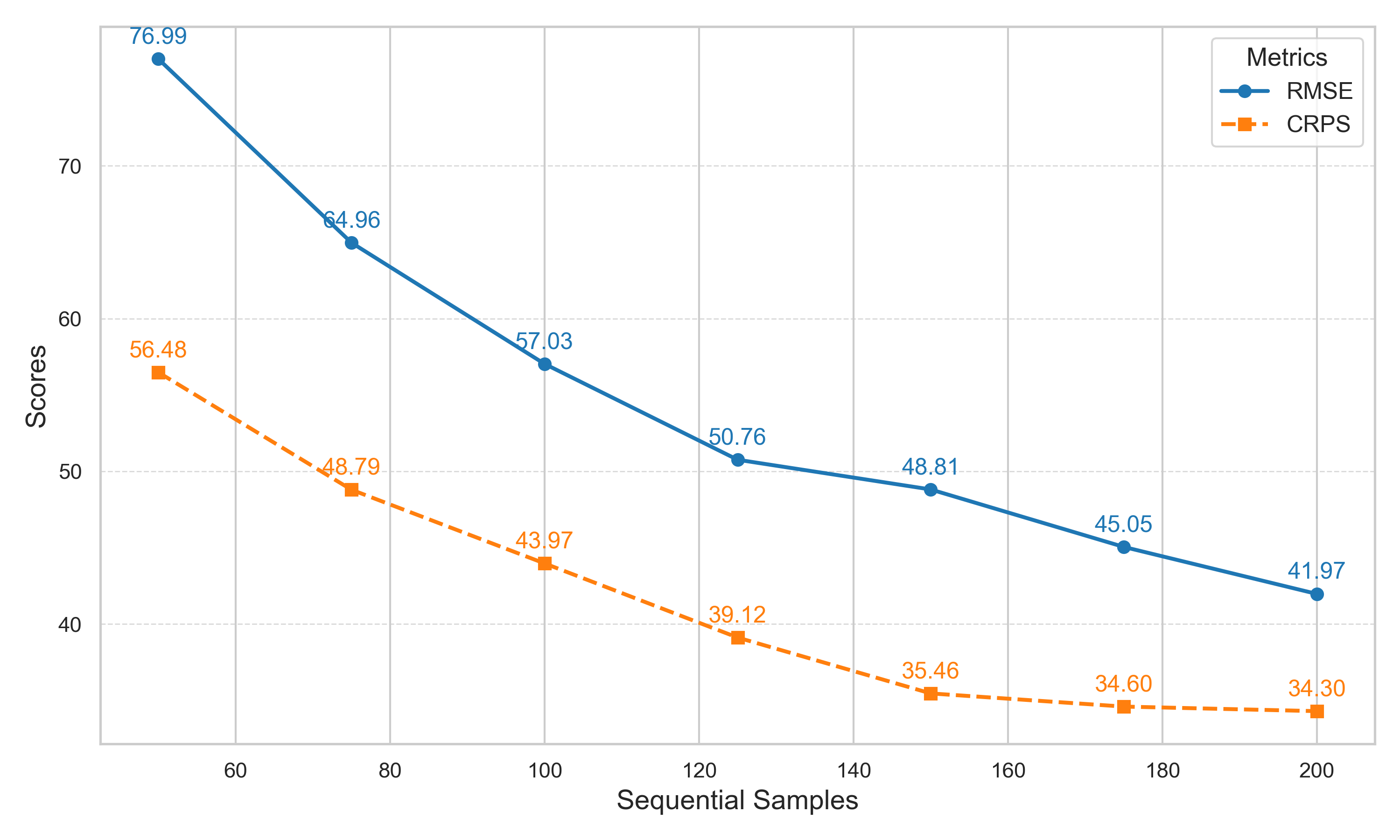}
    \caption{Effect of Sequential Sample Size on RMSE and CRPS for CA-SMART}
    \label{fig:sequential_sample_effect}
\end{figure}

\begin{table*}[!htb]
    \centering
    \small 
    \caption{Comparison of RMSE (Mean ± 95\% CI) for various machine learning models and the proposed CA-SMART approach across different training sample sizes.}
    \label{tab:ml_comparison}
    \begin{tabular}{lcccc}
        \hline
        \textbf{Approach} & \textbf{Sample Size = 100} & \textbf{Sample Size = 150} & \textbf{Sample Size = 200} & \textbf{Sample Size = 250} \\
        \hline
        Random Forest & 67.17 ± 4.15 & 60.52 ± 2.70 & 55.78 ± 1.92 & 51.48 ± 1.35 \\
        Gradient Boosting & 66.76 ± 4.16 & 57.13 ± 2.46 & 54.72 ± 2.36 & 50.97 ± 1.46 \\
        Extra Trees & 66.79 ± 3.13 & 60.46 ± 2.87 & 57.28 ± 1.70 & 53.94 ± 1.42 \\
        AdaBoost & 66.36 ± 4.22 & 58.03 ± 2.81 & 54.58 ± 2.09 & 51.55 ± 1.62 \\
        Lasso & 105.03 ± 5.08 & 107.84 ± 5.10 & 108.19 ± 4.50 & 104.54 ± 3.95 \\
        ElasticNet & 109.95 ± 5.00 & 112.05 ± 4.93 & 112.08 ± 4.50 & 108.46 ± 3.83 \\
        Bayesian Ridge & 85.17 ± 4.21 & 87.02 ± 3.74 & 85.44 ± 3.16 & 82.72 ± 2.82 \\
        Decision Tree & 66.77 ± 3.85 & 62.10 ± 3.16 & 57.61 ± 2.30 & 53.23 ± 1.86 \\
        KNN Regressor & 75.32 ± 3.89 & 68.30 ± 2.82 & 66.02 ± 1.96 & 62.89 ± 2.15 \\
        XGBoost & 95.95 ± 5.42 & 59.97 ± 2.71 & 56.58 ± 2.66 & 52.48 ± 1.69 \\
        MLP Regressor & 118.74 ± 5.11 & 122.88 ± 5.03 & 121.63 ± 4.72 & 115.51 ± 4.01 \\
        SVR & 86.41 ± 4.26 & 87.90 ± 3.64 & 86.47 ± 2.96 & 84.23 ± 2.64 \\
        CA-SMART & \textbf{64.96 ± 3.16} & \textbf{50.76 ± 3.74} & \textbf{45.05 ± 3.42} & \textbf{41.16 ± 3.18} \\
        \hline
    \end{tabular}
\end{table*}

\subsubsection{Effect of Sequential Sample Size on Model Performance}

In this section, we examine the impact of the number of sequential iterations on the performance of our CA-SMART framework. For this experiment, we fix the initial sample size at 25 and incrementally increased the number of sequential samples used to update the model. Figure~\ref{fig:sequential_sample_effect} shows how the RMSE and CRPS values evolve as the number of sequential samples is increased. As illustrated in Figure~\ref{fig:sequential_sample_effect}, both RMSE and CRPS scores exhibit a clear improvement as the number of sequential samples grows. In particular, the decrease in RMSE and CRPS is more rapid at the beginning, indicating that the model is able to select highly informative samples in the initial iterations. This early rapid improvement suggests that the CA-SMART framework effectively explores the design space by prioritizing the most informative points, thereby enhancing predictive performance. As the sequential sample size continues to increase, the rate of improvement in RMSE and CRPS gradually stabilizes. This indicates that, while additional samples continue to provide incremental gains, the model has already acquired substantial knowledge about the objective function and thus requires fewer new samples to achieve further improvements. This pattern highlights the efficiency of the CA-SMART approach in balancing exploration and exploitation, as it quickly gains an accurate approximation of the objective function with minimal sequential iterations.

\subsubsection{Comparison with Traditional Machine Learning Approaches}

In addition to comparing our sequential learning-based CA-SMART approach with other sequential methods (Shannon Surprise, Bayesian Surprise, and popular BO acquisition functions), we have also evaluated its performance against a range of traditional ML models that do not incorporate active learning. For a comprehensive assessment, we have calculated the mean and 95\% confidence intervals (CI) of the RMSE for different training sample sizes.  For the CA-SMART approach, the initial training sample size is fixed at 25 samples, and we vary the sequential sample size to reach different total sample counts. For the traditional ML models, we train each model on randomly sampled subsets of different sizes from the available training data. To ensure robustness and mitigate randomness, we have performed 30 independent runs for each configuration. Using 5-fold cross-validation with grid search, we have systematically evaluated combinations of hyperparameters across a diverse set of range for each ML model, selecting the set that maximized performance across folds.

The results, summarized in Table~\ref{tab:ml_comparison}, highlight that our CA-SMART approach achieves competitive performance compared to traditional ML models, often surpassing them. This demonstrates the advantage of CA-SMART's sequential, active learning approach, which can achieve higher predictive accuracy with fewer training samples by intelligently selecting the most informative samples. As shown in Table~\ref{tab:ml_comparison}, CA-SMART consistently achieves lower RMSE values with narrower confidence intervals across all sample sizes compared to traditional ML models. Traditional models generally require more samples to achieve similar performance, highlighting the effectiveness of CA-SMART's sequential learning process in selecting informative samples to optimize the learning process.

\section{Conclusion}

In this study, we introduce Confidence-Adjusted Surprise Measure for Active Resourceful Trials (CA-SMART), a novel active learning framework designed to optimize material property prediction while minimizing experimental trials. CA-SMART distinguishes itself through an adaptive and efficient strategy that dynamically balances exploration and exploitation within a Bayesian active learning framework. Unlike conventional acquisition functions in Bayesian Optimization, which rely on fixed criteria, CA-SMART integrates the Confidence-Adjusted Surprise (CAS) metric to prioritize observations based on their potential to refine model understanding. 
CAS refines traditional surprise-based learning by incorporating flat prior comparison and confidence adjustments, distinguishing it from Shannon and Bayesian surprise metrics. Shannon surprise, which relies solely on entropy, often amplifies uncertainty without differentiating between noise and true information gain. Bayesian surprise, though accounting for prior beliefs, is prone to model biases and can inefficiently allocate resources by over-prioritizing unexpected data points. CAS overcomes these limitations by introducing a flat prior for unbiased surprise estimation and incorporating confidence adjustments to moderate sampling in high-uncertainty or low-likelihood regions. This ensures that CA-SMART prioritizes observations that are both surprising and reliably informative, leading to more efficient resource allocation. CA-SMART’s sampling strategy further enhances its effectiveness. It initializes with Sobol sequence sampling, ensuring a well-distributed and unbiased representation of the design space. As learning progresses, it dynamically balances exploration and exploitation by selecting under-sampled regions to enhance diversity while refining local response surfaces through adaptive sampling. This prevents excessive focus on highly uncertain regions, which can lead to inefficient data collection, and instead directs experimental efforts toward observations that meaningfully improve model predictions.

CA-SMART is evaluated on benchmark synthetic functions and a real-world case study on material property prediction. The results demonstrate that CA-SMART consistently outperforms both traditional BO acquisition functions, surprise-based methods and traditional ML approaches. Its ability to dynamically guide sample selection enables efficient reduction in RMSE while achieving faster convergence. The CRPS results further validate its effectiveness, showing improved prediction accuracy and reduced uncertainty compared to competing approaches. Additionally, an evaluation of the impact of initial and sequential sample sizes has been conducted to help future practitioners determine the optimal number of samples required for efficient model training. 

CA-SMART establishes itself as a robust and adaptive framework for active learning in resource-constrained environments. By intelligently balancing global exploration and local refinement, it accelerates convergence and enables accurate approximation of complex response surfaces. Our work lays the foundation for further exploration of surprise-based metrics in active learning frameworks. The versatility of CA-SMART suggests that it could be adapted to other domains where active learning and resource efficiency are critical, including chemistry, physics, and biomedical research. For practitioners and researchers, CA-SMART offers a powerful tool for accelerating the material discovery process through an autonomous manufacturing platform, providing a reliable and efficient method for identifying and exploring promising material properties.

\section*{CRediT Authorship Contribution Statement}

\textbf{Ahmed Shoyeb Raihan:} Conceptualization, Data curation, Formal analysis, Investigation, Methodology, Software, Validation, Visualization, Writing – original draft. \textbf{Zhichao Liu:} Validation, Supervision, Writing – review and editing. \textbf{Tanveer Hossain Bhuiyan:} Validation, Supervision, Writing – review and editing. \textbf{Imtiaz Ahmed:} Conceptualization, Formal analysis, Investigation, Resources, Validation,  Project administration, Supervision, Writing – review and editing.

\section*{Declaration of Competing Interest}

The authors declare that they have no known competing financial interests or personal relationships that could have appeared to influence the work reported in this paper.

\section*{Data Availability}

Data will be made available on request.

\section*{Acknowledgements}

The authors would like to thank the Department of Industrial and Management Systems Engineering (IMSE), West Virginia University (WVU) for their invaluable support.









\bibliographystyle{elsarticle-num} 
\bibliography{main}

\end{document}